\documentclass[11pt]{article}

\usepackage[final]{acl}

\usepackage{times}
\usepackage{latexsym}

\usepackage[T1]{fontenc}
\usepackage[utf8]{inputenc}

\usepackage{microtype}

\usepackage{inconsolata}

\usepackage{graphicx}

\usepackage{minitoc}
\usepackage{amsmath}
\usepackage{amssymb}
\usepackage{mathtools}
\usepackage{amsthm}
\usepackage{multirow}
\usepackage{xcolor}
\usepackage{colortbl}
\usepackage{booktabs}

\usepackage{algorithm}
\usepackage{algorithmicx}
\usepackage{algpseudocode}
\usepackage{verbatim}
\usepackage{subcaption}
\usepackage{fancyvrb}
\usepackage{fvextra}
\usepackage{xurl}
\usepackage{times}
\usepackage{helvet}
\usepackage{courier}
\usepackage[most]{tcolorbox}
\usepackage{wrapfig}
\usepackage{longtable}
\usepackage{afterpage}
\usepackage{subcaption}
\usepackage{algorithm}
\usepackage{algpseudocode}
\usepackage{array}
\usepackage{pifont}

\definecolor{GREEN}{HTML}{62b197}
\definecolor{RED}{HTML}{e18e6d}

\DeclareUnicodeCharacter{FF5C}{\textbar}

\definecolor{DarkGreen}{RGB}{0,120,0}
\definecolor{DarkRed}{RGB}{180,40,40}
\newcommand{\cmark}{\textcolor{DarkGreen}{\ding{51}}}
\newcommand{\xmark}{\textcolor{DarkRed}{\ding{55}}}

\newcommand{\ms}[2]{#1{{\scriptsize#2}}}

\title{CCTU: A Benchmark for Tool Use under Complex Constraints}

\author{%
Junjie Ye$^{1}$\thanks{Equal contribution.}, Guoqiang Zhang$^{1*}$, Wenjie Fu$^{1*}$, Zelin Li$^{2}$,\\
\textbf{Tao Gui$^{1}$, Qi Zhang$^{1}$, Xuanjing Huang$^{1}$\thanks{Corresponding author.}}\\
\\
$^1$Fudan University $^2$Northwestern University\\
\\
\texttt{jjye23@m.fudan.edu.cn, xjhuang@fudan.edu.cn}\\
}

\begin{document}
\maketitle
\begin{abstract}
Solving problems through tool use under explicit constraints constitutes a highly challenging yet unavoidable scenario for large language models (LLMs), requiring capabilities such as function calling, instruction following, and self-refinement. However, progress has been hindered by the absence of dedicated evaluations. To address this, we introduce \emph{CCTU}, a benchmark for evaluating LLM tool use under complex constraints. CCTU is grounded in a taxonomy of 12 constraint categories spanning four dimensions (i.e., resource, behavior, toolset, and response). The benchmark comprises 200 carefully curated and challenging test cases across diverse tool-use scenarios, each involving an average of seven constraint types and an average prompt length exceeding 4,700 tokens. To enable reliable evaluation, we develop an executable constraint validation module that performs step-level validation and enforces compliance during multi-turn interactions between models and their environments. We evaluate nine state-of-the-art LLMs in both thinking and non-thinking modes. Results indicate that when strict adherence to all constraints is required, no model achieves a task completion rate above 20\%. Further analysis reveals that models violate constraints in over 50\% of cases, particularly in the resource and response dimensions. Moreover, LLMs demonstrate limited capacity for self-refinement even after receiving detailed feedback on constraint violations, highlighting a critical bottleneck in the development of robust tool-use agents. To facilitate future research, we release the data\footnote{\url{https://huggingface.co/datasets/Junjie-Ye/CCTU}} and code\footnote{\url{https://github.com/Junjie-Ye/CCTU}}.
\end{abstract}

\section{Introduction}
\label{sec:intro}

\begin{figure*}[!t]
    \centering
    \includegraphics[width=\linewidth]{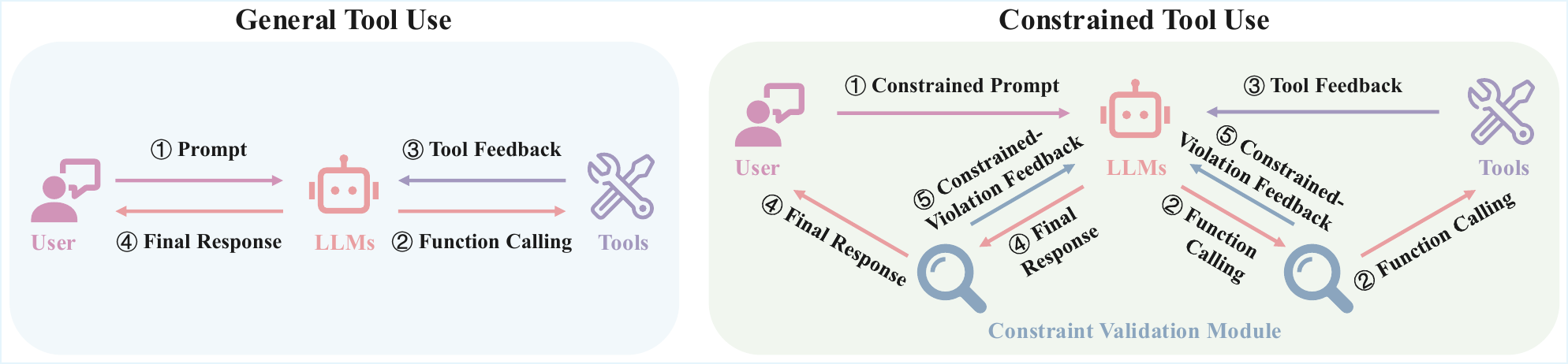}
    \caption{Comparison between general tool use and constrained tool use. The constrained setting introduces a constraint validation module that performs step-level verification across multi-turn interactions and provides feedback when violations occur.}
    \label{fig:ctu}
    % \vspace{-4mm}
\end{figure*}

Solving problems through tool use under explicit constraints poses a significant challenge for large language models (LLMs)~\citep{Claude4.6, Gemini3, DeepSeek-R1, GPT5.2}. As illustrated in Figure~\ref{fig:ctu}, such scenarios require models to demonstrate strong function-calling abilities~\citep{tool-survey} for accurate tool selection and invocation, reliable instruction-following skills~\citep{IF-survey} to consistently adhere to specified constraints throughout the process, and effective self-refinement mechanisms~\citep{Self-Refine} to adapt their behavior during dynamic interactions. At the same time, such requirements are unavoidable in practical deployments. For instance, LLMs must operate under constraints such as latency limits~\citep{budget}, restrictions on tool access frequency~\citep{Toolllm}, and predefined response formatting rules~\citep{AGENTIF} when using external tools.

Existing studies evaluate specific aspects of model capability in isolation. One line examines tool selection and invocation across diverse settings, from single-turn~\citep{Toolformer, ToolQA} to multi-turn~\citep{Toolllm, ACEBench} and complex scenarios~\citep{FamilyTool, ToolHop}. Another focuses on instruction-following, covering rule-verifiable~\citep{IFEval} and nuanced dimensions~\citep{MultiChallenge, SysBench}. A growing body of work also explores self-refinement~\citep{Self-Refine, Reflexion}.

However, these benchmarks do not capture integrated performance in constrained tool-use scenarios. A model that correctly invokes tools may still fail to adhere to constraints, while one with strong instruction-following may struggle to differentiate tool roles. Whether models can self-refine after constraint violations in dynamic settings remains underexplored, highlighting the need for benchmarks that systematically assess constrained tool-use performance.

To address this, we introduce CCTU, a benchmark for evaluating LLM tool use under complex constraints. We develop a taxonomy of 12 constraint categories across four dimensions (i.e., resource, behavior, toolset, and response) and curate 200 challenging test cases covering diverse tool-use scenarios. Through LLM-based filtering and manual verification, each finalized case involves an average of seven constraint types with prompt lengths exceeding 4,700 tokens. We additionally develop an executable constraint validation module for step-level compliance checking during multi-turn interactions.

We evaluate nine state-of-the-art LLMs on CCTU in both thinking and non-thinking modes. The best-performing model achieves less than 20\% task completion rate under strict constraint adherence, with most falling below 15\%, highlighting severe limitations in integrated capabilities. Further analysis reveals that models violate constraints in over 50\% of cases, particularly in the resource and response dimensions, and struggle to self-refine based on detailed violation feedback, which is a significant bottleneck for robust tool-use agents.

Our contributions are summarized as follows:
\begin{itemize} 
    \item We introduce CCTU, a benchmark comprising 200 test cases with 12 constraint categories across four dimensions, designed to evaluate tool use under complex constraints. 
    \item We design a constraint validation module that enables step-level supervision and precise evaluation of complete interaction trajectories. 
    \item We conduct comprehensive evaluations of nine state-of-the-art LLMs under both thinking and non-thinking modes, revealing substantial limitations in constrained tool use and identifying key error patterns and self-refinement bottlenecks.
\end{itemize}

\begin{table*}[!t]
\centering
\resizebox{\linewidth}{!}
{
\begin{tabular}{lccccccccccc}
\toprule
\multirow{2}{*}{\textbf{Benchmarks}} & \multicolumn{4}{c}{\textbf{Basic Information}} & \multicolumn{4}{c}{\textbf{Constraint Dimensions}} & \multicolumn{3}{c}{\textbf{Evaluated Capabilities}} \\
\cmidrule(lr){2-5} \cmidrule(lr){6-9} \cmidrule(lr){10-12}
 & \textbf{\textit{\#Number}} & \textbf{\textit{Avg. Tools}} & \textbf{\textit{Avg. Len.}} & \textbf{\textit{Precise Eval.}} 
 & \textbf{\textit{Resource}} & \textbf{\textit{Behavior}} & \textbf{\textit{Toolset}} & \textbf{\textit{Response}} 
 & \textbf{\textit{Func. Call.}} & \textbf{\textit{Ins. Follow.}} & \textbf{\textit{Self-Refine.}} \\
\midrule
IFEval~\citep{IFEval}        & 541  & 0     & 54   & \cmark & \xmark & \xmark & \xmark & \cmark & \xmark & \cmark & \xmark \\
IFBench~\citep{IF-Bench}       & 300  & 0     & 90   & \cmark & \xmark & \xmark & \xmark & \cmark & \xmark & \cmark & \xmark \\
MultiChallenge~\citep{MultiChallenge} & 273  & 0     & 1636 & \xmark     & \xmark & \xmark & \xmark & \cmark & \xmark & \cmark & \xmark \\
BFCL v4~\citep{BFCL}       & 5088 & 7.08  & 1446 & \cmark & \xmark & \xmark & \cmark & \cmark & \cmark & \cmark & \xmark \\
$\tau$-bench~\citep{tau-bench}  & 165  & 12.36 & 4245 & \cmark & \xmark & \xmark & \cmark & \xmark     & \cmark & \xmark     & \xmark \\
FTRL~\citep{FTRL}          & 200  & 9.26  & 3864 & \cmark & \xmark & \xmark & \cmark & \xmark     & \cmark & \xmark     & \xmark \\
% SOPBench~\citep{SOPBench}      & 903  & 23.7  & 5332 & \cmark & \xmark & \cmark & \cmark & \xmark & \cmark & \cmark & \xmark \\
AGENTIF~\citep{AGENTIF}       & 707  & 3.92  & 2387 & \xmark     & \xmark & \cmark & \cmark & \cmark & \cmark & \cmark & \xmark \\
\rowcolor{gray!10} CCTU (Ours)   & 200  & 9.26  & 4754 & \cmark & \cmark & \cmark & \cmark & \cmark & \cmark & \cmark & \cmark \\
\bottomrule
\end{tabular}
}
\caption{Comparison of benchmarks across basic information, constraint dimensions, and evaluated capabilities.}
\label{tab:compare}
% \vspace{-4mm}
\end{table*}

\section{Related Work}
\paragraph{Evaluations for Tool Use}
Tool use has become a core LLM application, spurring extensive evaluation research spanning diverse scenarios~\citep{TEval, Toolalpaca} and evolving toward complex settings such as multi-hop and parallel tasks~\citep{BFCL, ToolHop}, reflecting the broader trend toward production-oriented tasks~\citep{augmented-survey, AppWorld, agent-survey}. However, most prior work primarily evaluates whether models eventually solve user queries, with limited consideration of constraints governing tool use. In contrast, our work evaluates tool use under complex constraints, emphasizing whether models can plan action trajectories in accordance with specified restrictions and systematically analyzing how different constraint types affect performance.

\paragraph{Evaluations for Instruction Following}
A substantial body of work evaluates instruction-following capabilities of LLMs. Early studies relied on template-based methods~\citep{IF-Bench, COLLIE, IFEval}, while more advanced approaches increased instruction complexity with LLM-as-a-judge paradigms~\citep{MultiChallenge, SysBench, InverseIF}. Recent work has extended such evaluations to agentic settings~\citep{AGENTIF}. However, these studies primarily assess whether responses violate constraints in static instructions. In contrast, we develop an executable constraint validation module that conducts step-level compliance checks during multi-turn interactions.

% \paragraph{Self-Refinement of LLMs}
% Self-refinement is a critical capability of LLMs, referring to their ability to enhance problem-solving through iterative updates to their own outputs~\citep{Self-Refine, Reflexion}. It was initially investigated in reasoning-intensive tasks~\citep{DeepSeekMath, CoT}. Subsequent work extended this idea to tool-use scenarios, enabling models to incorporate environmental feedback when planning subsequent actions~\citep{tool-learning, TaskBench, React}. However, whether models can effectively self-refine when their actions violate explicit constraints remains unexplored. In this paper, when models violate constraints during tool use, we provide detailed feedback through a constraint validation module. By leveraging executable validation mechanisms, we systematically evaluate whether models can recognize and correct their violations, thereby assessing their self-refinement capability in constrained multi-turn interactions.

\section{CCTU}
% In this section, we present a detailed overview of CCTU. Section~\ref{sec:taxonomy} introduces the constraint taxonomy underlying its design, Section~\ref{sec:construction} describes the construction process, and Section~\ref{sec:analysis} provides a comprehensive analysis of the dataset.

\subsection{Constraint Taxonomy}
\label{sec:taxonomy}

Derived from practical application requirements, we identify 12 representative constraints to enable precise evaluation in tool-use scenarios. Organized into four dimensions, these constraints form a structured taxonomy that underpins the construction of diverse and challenging test cases.

% \paragraph{Resource}
\textbf{Resource} constraints stem from the dual requirements of efficiency and quality, placing stringent demands on global planning to avoid both insufficient utilization and excessive trial-and-error.
% \begin{itemize}
%     \item 
    1) \textbf{Interaction rounds} limit the total number of exchanges between the model and the environment, requiring the model to produce a final response within the specified bound. Exceeding this limit results in automatic task termination.
    % \item 
    2) \textbf{Tool call count} restricts the total number of tool invocations permitted during task execution. Any invocation attempt beyond this upper bound is disregarded.
    % \item
    3) \textbf{Specific tool call count} constrains the number of times designated tools may be invoked, emphasizing the need for deliberate planning and efficient allocation of these tools. Exceeding the limit renders these tools unavailable, while other tools remain accessible.
% \end{itemize}

% \paragraph{Behavior}
\textbf{Behavior} constraints arise from the need to maintain controllability, requiring models to follow predefined behavior norms. Although restricting the decision space, they also provide structural guidance for effective task execution.
% \begin{itemize}
    % \item 
    1) \textbf{Sequential dependencies} govern the order of tool invocations, often as conditional requirements. For instance, a model may be required to obtain authorization before accessing certain data. Invocations that violate these dependencies are rejected, and feedback indicates which preceding tools must be invoked.
    % \item
    2) \textbf{Parallel dependencies} define conditional relationships between concurrently invoked tools. For instance, a model may be required to log data while updating it. Violations of parallel dependencies are similarly rejected, with feedback provided to guide the model.
    % \item 
    3) \textbf{Parallel calls count} constrains the allowable range of parallel tool calls during task execution, requiring the model to correctly decompose complex intentions and distinguish unrelated subtasks. Parallel calls exceeding the upper limit are ignored, while fewer calls than the lower limit prevent the model from proactively completing the task.
% \end{itemize}

% \paragraph{Toolset}
\textbf{Toolset} constraints are fundamental to tool-use scenarios. They define the characteristics and usage specifications of tools through structured documentation. While previous work often relied on tool execution outcomes to implicitly enforce these constraints, we perform explicit validations.
% \begin{itemize}
    % \item 
    1) \textbf{Available tools and parameters} restrict the set of tools that the model is permitted to invoke, as well as the allowable parameter ranges. Any invocation beyond this predefined scope is considered a hallucinated call.
    % \item 
    2) \textbf{Required parameters} define the mandatory arguments that must be provided when invoking a tool. Omission of any required parameter results in invocation failure.
    % \item 
    3) \textbf{Parameter types} require the model to correctly identify parameter value formats and perform appropriate type conversions when necessary. Supplying a value of an incorrect type results in invocation failure.
% \end{itemize}

% \paragraph{Response}
\textbf{Response} constraints stem from requirements concerning the form and structure of model outputs, mandating that final responses adhere to predefined specifications. Responses that violate any constraint must be regenerated.
% \begin{itemize}
    % \item 
    1) \textbf{Length} restricts the allowable range of the model’s final response.
    % \item 
    2) \textbf{Format} specifies the presentation style of the final response, such as plain text, JSON, or tabular representations.
    % \item 
    3) \textbf{Content} imposes specific requirements on elements that must appear in the final response, including designated languages, identifiers, keywords, or other prescribed information.
% \end{itemize}

\begin{figure}[!t]
    \centering
    \includegraphics[width=\linewidth]{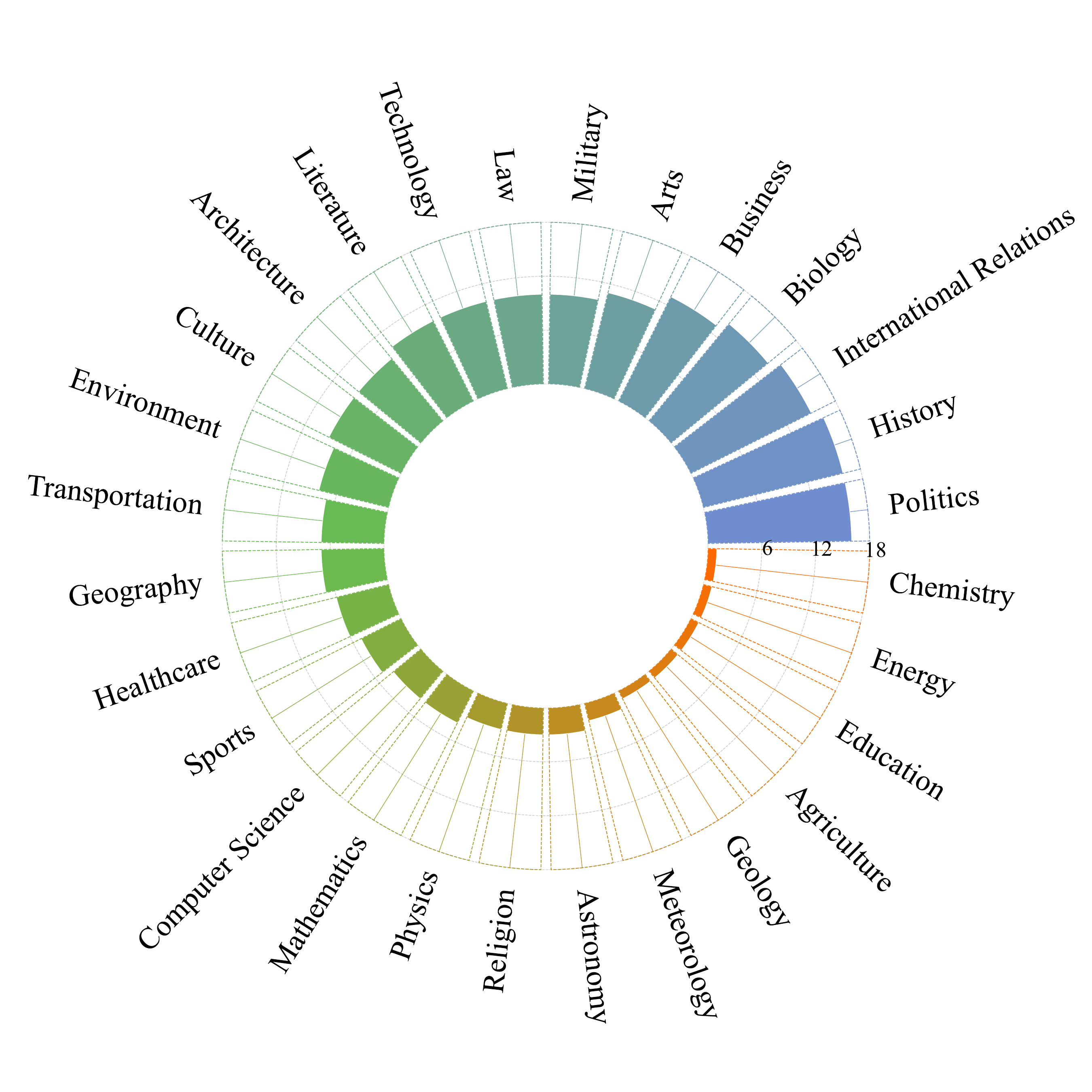}
    \caption{Distribution of CCTU queries across 28 domains.}
    \label{fig:domain}
    % \vspace{-4mm}
\end{figure}

\subsection{Benchmark Construction}
\label{sec:construction}

We construct 200 test cases through a systematic pipeline\footnote{We summarize the pipeline in Appendix~\ref{sec:pipeline}.} comprising four components: prompt sourcing, constraint integration, executable constraint validation, and quality control.\footnote{Prompts used in the pipeline are provided in Appendix~\ref{sec:prompt}.}

\paragraph{Prompt Sourcing}
We adopt FTRL~\citep{FTRL} as our initial dataset, which comprises four categories based on subquery interrelationships: single-hop, parallel single-hop, multi-hop, and parallel multi-hop, with 50 instances each. Each instance specifies the subqueries, required tools, and expected answers, enabling straightforward verification of task completion. Moreover, each instance involves an average of 9.26 locally executable tools without explicit constraints, placing substantial demands on function-calling capabilities while providing a flexible foundation for incorporating various constraints.

\begin{figure}[!t]
    \centering
    \includegraphics[width=\linewidth]{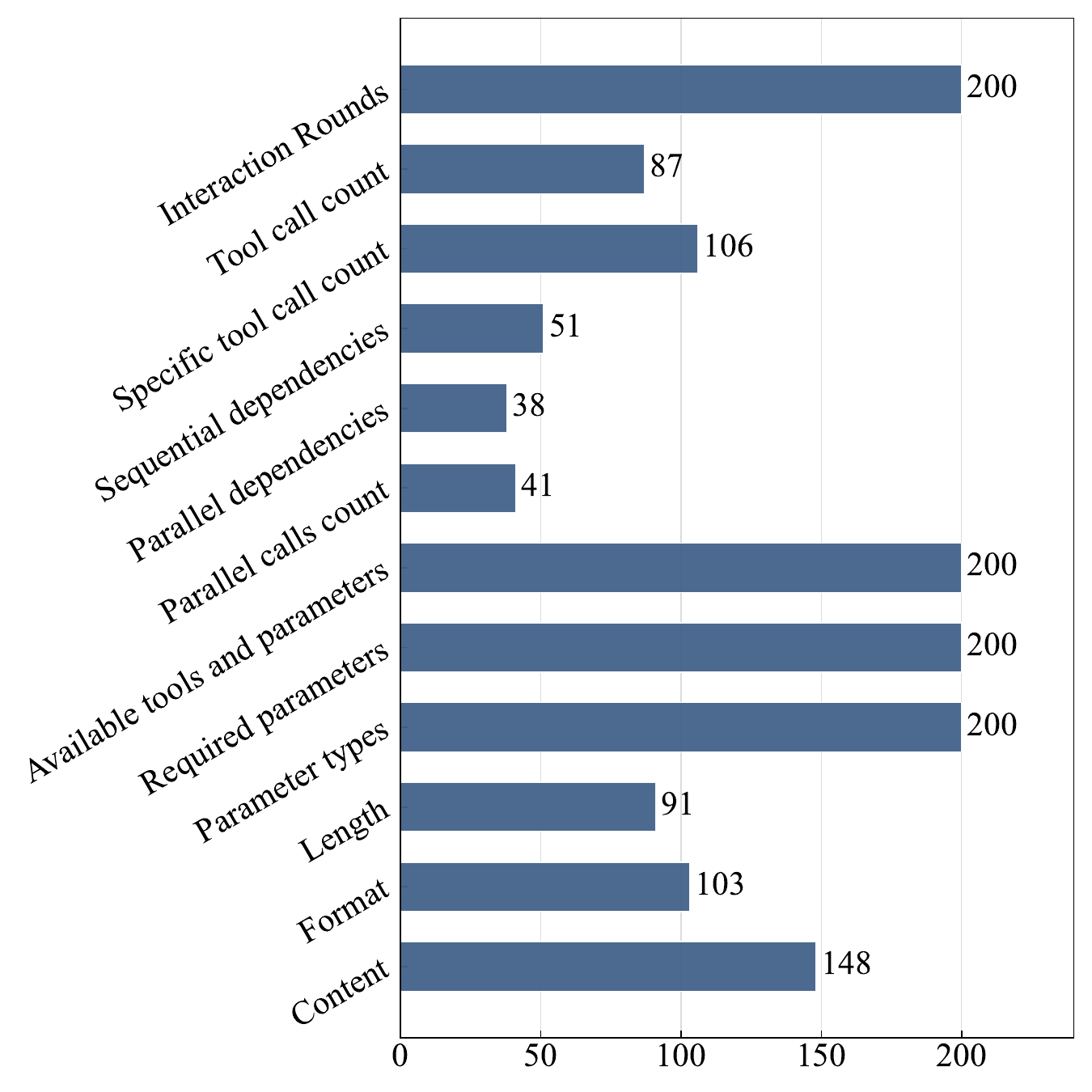}
    \caption{The number of samples associated with each constraint category in CCTU.}
    \label{fig:constype}
    % \vspace{-4mm}
\end{figure}

\paragraph{Constraint Integration}
To integrate our constraints into the initial dataset, we design an automated workflow that rewrites existing instances in an efficient and controllable manner. The workflow consists of four stages. \textbf{1) Reference trajectory generation.} To avoid unrealistic settings or logical contradictions, we first use LLMs to sample a correct solution trajectory for each instance as a reference.\footnote{We employ Qwen3-32B~\citep{Qwen3} in our pipeline due to its strong performance at low computational cost.} To improve sampling effectiveness given the dataset's difficulty~\citep{FTRL}, we provide the model with unsolved subqueries and local tool implementations. The resulting trajectories intentionally retain trial-and-error steps to increase diversity during subsequent constraint expansion. \textbf{2) Controlled constraint expansion.} For each instance, we iteratively introduce constraints using LLMs, iterating over constraint types except those in the Toolset dimension\footnote{Constraints in the Toolset dimension are introduced through tool documents in the original dataset.} with a 50\% selection probability per type. The model incorporates each constraint consistently with the reference trajectory. We impose structural rules: sequential dependencies are excluded from single-hop and parallel single-hop instances, and parallel dependencies and parallel call count constraints from single-hop and multi-hop settings. \textbf{3) LLM-based filtering.} We employ LLMs to verify the consistency and feasibility of modified instances, identifying conflicts among constraints and ensuring alignment with scenario structures (e.g., flagging an interaction round limit of one in a multi-hop scenario). If inconsistencies are detected, the process returns to the previous stage until verification succeeds. \textbf{4) Task context integration.} Since the original dataset contains only user queries, we use LLMs to generate scenario-level task contexts that provide constraint-independent background descriptions, combined with constrained specifications to form complete use cases.

\paragraph{Constraint Validation}
We design a constraint validation module that operates after each model output step (Figure~\ref{fig:ctu}), evaluating whether the output satisfies predefined constraints. If compliant, it triggers corresponding tool invocations or concludes the workflow without altering execution logic. If a violation is detected, it returns detailed feedback and prompts revision. This feedback is injected as tool or user messages, preserving the model's original inference configuration. The module uses LLM-pre-generated executable validation code for each constraint, which analyzes accumulated interaction logs to check compliance.

\begin{figure}
    \centering
    \includegraphics[width=\linewidth]{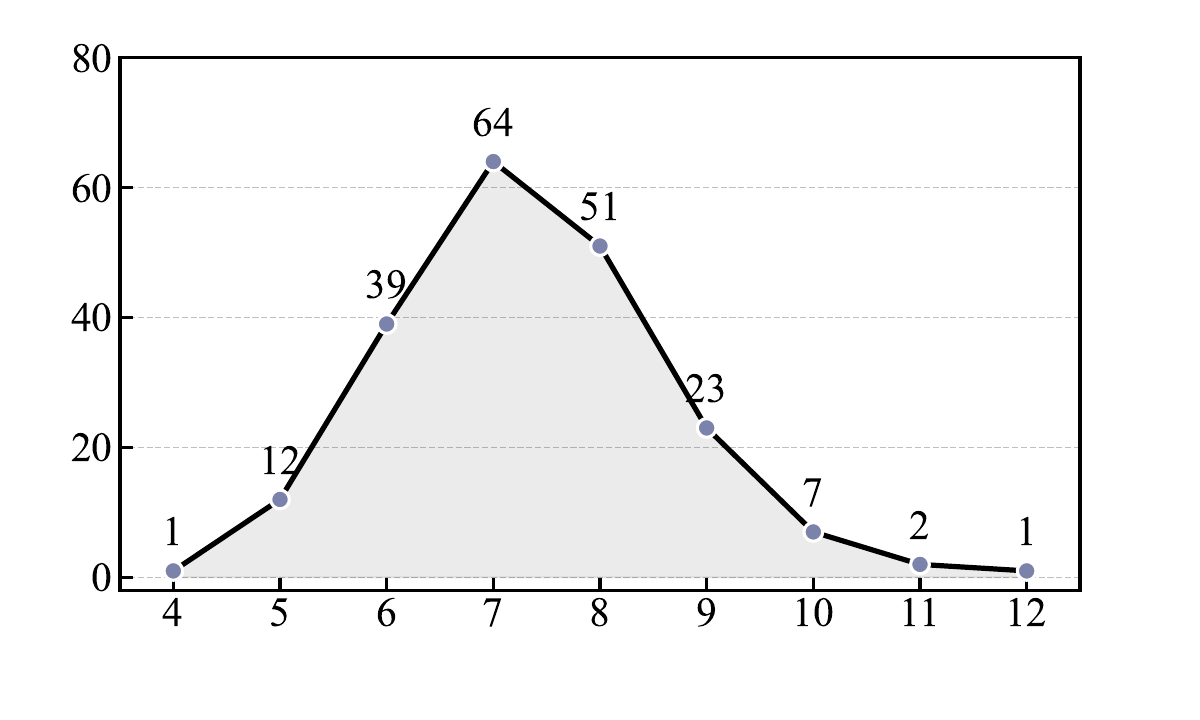}
    \caption{Distribution of the number of constraint categories per sample in CCTU.}
    \label{fig:consnum}
    % \vspace{-4mm}
\end{figure}

\paragraph{Quality Control}
To ensure data quality, we manually verify each data instance and its constraint validation code. \textbf{1) Data verification.} Each instance is reviewed by a computer science graduate student for conflicting constraints, unreasonable settings, and logical inconsistencies, with manual revision as needed. A second student then evaluates the instance; the process terminates only when two consecutive annotators agree it is free of issues, otherwise re-entering revision until consensus. \textbf{2) Code verification.} The same workflow is applied to inspect constraint validation code, concluding when two consecutive annotators confirm correctness.\footnote{More details for the process are provided in Appendix~\ref{sec:annotatation}.}

\begin{figure}
    \centering
    \includegraphics[width=\linewidth]{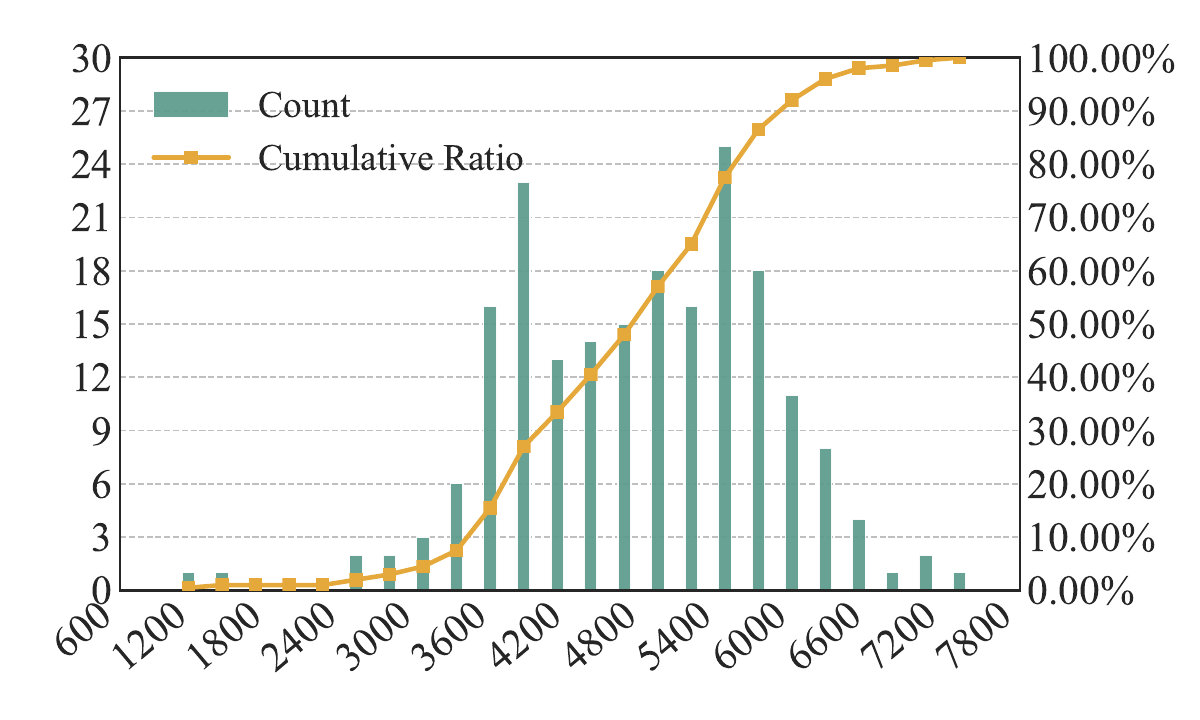}
    \caption{Distribution of data lengths in CCTU, measured using the Qwen3 tokenizer.}
    \label{fig:length}
    % \vspace{-4mm}
\end{figure}

\subsection{Data Analysis}
\label{sec:analysis}

We analyze CCTU along four key characteristics. Table~\ref{tab:compare} compares CCTU with existing benchmarks.

\paragraph{Diverse Domains}
CCTU is built upon FTRL and covers four categories of compositional relationships among subqueries, enabling the evaluation across diverse scenarios. Furthermore, we categorize the domains represented in the dataset. As shown in Figure~\ref{fig:domain}, CCTU spans 28 distinct domains, including specialized fields such as politics and sports, as well as everyday domains such as culture and tourism. This breadth ensures comprehensive evaluation of model performance across varied contexts, enhancing both its representativeness and practical relevance.

\paragraph{Complex Constraints}
Figure~\ref{fig:constype} presents the number of instances associated with each constraint type. Constraints in the behavior dimension appear less frequently due to their dependence on specific scenario structures, whereas constraints in the other three dimensions are present in the majority of the dataset. Every instance simultaneously includes constraints from both the resource and toolset dimensions. Figure~\ref{fig:consnum} further shows that each data point contains between 4 and 12 constraint types, with an average of 7 constraints per instance. This design highlights the diversity and complexity of constraint combinations within CCTU.

\begin{table*}[t]
\centering
\resizebox{\linewidth}{!}
{
\begin{tabular}{lcccccccccc}
\toprule
\multirow{2}*{\textbf{Models}} 
& \multicolumn{2}{c}{\textbf{Single-Hop}} 
& \multicolumn{2}{c}{\textbf{Parallel Single-Hop}} 
& \multicolumn{2}{c}{\textbf{Multi-Hop}} 
& \multicolumn{2}{c}{\textbf{Parallel Multi-Hop}} 
& \multicolumn{2}{c}{\textbf{Overall}} \\
\cmidrule(lr){2-3} \cmidrule(lr){4-5} \cmidrule(lr){6-7} \cmidrule(lr){8-9} \cmidrule(lr){10-11}
& \textit{\textbf{SR}} & \textit{\textbf{PSR}} & \textit{\textbf{SR}} & \textit{\textbf{PSR}} & \textit{\textbf{SR}} & \textit{\textbf{PSR}} & \textit{\textbf{SR}} & \textit{\textbf{PSR}} & \textit{\textbf{SR}} & \textit{\textbf{PSR}} \\
\midrule
\rowcolor{gray!10} \multicolumn{11}{c}{\textit{Thinking Mode}} \\

GPT-5.2 
& \ms{32.67}{1.89} & \ms{\textbf{24.67}}{3.40}
& \ms{24.67}{1.89} & \ms{\textbf{17.33}}{0.94}
& \ms{25.33}{2.49} & \ms{20.67}{2.49}
& \ms{15.33}{0.94} & \ms{10.00}{1.63}
& \ms{24.50}{0.82} & \ms{\textbf{18.17}}{1.03} \\

GPT-5.1 
& \ms{25.33}{5.25} & \ms{20.00}{7.12}
& \ms{20.67}{2.49} & \ms{16.00}{0.00}
& \ms{22.67}{1.89} & \ms{20.67}{1.89}
& \ms{22.67}{0.94} & \ms{9.33}{2.49}
& \ms{22.83}{2.01} & \ms{16.50}{2.83} \\

Claude Opus 4.6 
& \ms{\textbf{34.67}}{4.99} & \ms{10.00}{1.63}
& \ms{\textbf{30.67}}{2.49} & \ms{13.33}{0.94}
& \ms{\textbf{38.67}}{3.40} & \ms{\textbf{23.33}}{1.89}
& \ms{\textbf{32.67}}{1.89} & \ms{\textbf{12.67}}{0.94}
& \ms{\textbf{34.17}}{2.25} & \ms{14.83}{1.03} \\

Seed-2.0-Pro 
& \ms{22.67}{3.77} & \ms{19.33}{4.99}
& \ms{20.67}{5.25} & \ms{12.67}{3.40}
& \ms{22.67}{0.94} & \ms{18.67}{0.94}
& \ms{15.33}{1.89} & \ms{8.67}{2.49}
& \ms{20.33}{2.62} & \ms{14.83}{2.87} \\

Qwen3.5-Plus 
& \ms{20.67}{4.99} & \ms{5.33}{2.49}
& \ms{23.33}{0.94} & \ms{8.00}{1.63}
& \ms{32.00}{2.83} & \ms{21.33}{0.94}
& \ms{23.33}{1.89} & \ms{8.00}{1.63}
& \ms{24.83}{1.03} & \ms{10.67}{1.55} \\

Gemini 3 Pro 
& \ms{23.33}{2.49} & \ms{12.00}{3.27}
& \ms{28.00}{1.63} & \ms{16.00}{1.63}
& \ms{14.67}{0.94} & \ms{11.33}{2.49}
& \ms{11.33}{3.40} & \ms{2.67}{0.94}
& \ms{19.33}{1.84} & \ms{10.50}{1.08} \\

DeepSeek-V3.2 
& \ms{15.33}{1.89} & \ms{6.67}{2.49}
& \ms{22.67}{1.89} & \ms{12.00}{1.63}
& \ms{26.00}{4.32} & \ms{16.67}{3.40}
& \ms{8.00}{1.63} & \ms{0.67}{0.94}
& \ms{18.00}{1.08} & \ms{9.00}{0.41} \\

OpenAI o3 
& \ms{22.67}{1.89} & \ms{17.33}{1.89}
& \ms{7.33}{0.94} & \ms{4.00}{1.63}
& \ms{13.33}{0.94} & \ms{10.00}{1.63}
& \ms{4.00}{1.63} & \ms{1.33}{0.94}
& \ms{11.83}{1.31} & \ms{8.17}{1.03} \\

Kimi K2.5 
& \ms{22.67}{3.77} & \ms{4.67}{2.49}
& \ms{26.00}{3.27} & \ms{10.67}{3.77}
& \ms{20.00}{2.83} & \ms{10.67}{2.49}
& \ms{16.67}{4.11} & \ms{4.67}{2.49}
& \ms{21.33}{2.05} & \ms{7.67}{1.65} \\

\midrule
\rowcolor{gray!10} \multicolumn{11}{c}{\textit{Non-Thinking Mode}} \\

GPT-5.2 
& \ms{28.00}{3.27} & \ms{\textbf{24.00}}{1.63}
& \ms{19.33}{0.94} & \ms{15.33}{2.49}
& \ms{17.33}{4.11} & \ms{14.00}{4.90}
& \ms{16.67}{0.94} & \ms{10.67}{0.94}
& \ms{20.33}{0.62} & \ms{\textbf{16.00}}{1.87} \\

Claude Opus 4.6 
& \ms{\textbf{38.00}}{1.63} & \ms{12.00}{0.00}
& \ms{\textbf{29.33}}{0.94} & \ms{13.33}{0.94}
& \ms{\textbf{38.00}}{2.83} & \ms{\textbf{23.33}}{1.89}
& \ms{\textbf{32.67}}{0.94} & \ms{\textbf{13.33}}{0.94}
& \ms{\textbf{34.50}}{1.08} & \ms{15.50}{0.41} \\

GPT-5.1 
& \ms{22.67}{2.49} & \ms{19.33}{0.94}
& \ms{19.33}{2.49} & \ms{\textbf{16.67}}{1.89}
& \ms{16.67}{2.49} & \ms{14.00}{1.63}
& \ms{14.00}{0.00} & \ms{6.67}{3.40}
& \ms{18.17}{0.85} & \ms{14.17}{0.62} \\

Kimi K2.5 & 
\ms{19.33}{3.40} & \ms{6.67}{0.94} & \ms{29.33}{0.94} & \ms{14.00}{0.00} & \ms{25.33}{1.89} & \ms{15.33}{0.94} & \ms{16.67}{0.94} & \ms{6.67}{0.94} & \ms{22.67}{0.85} & \ms{10.67}{0.62} \\

Gemini 3 Pro & 
\ms{22.67}{6.60} & \ms{12.67}{4.71} & \ms{26.67}{2.49} & \ms{14.67}{0.94} & \ms{16.00}{3.27} & \ms{11.33}{0.94} & \ms{10.67}{1.89} & \ms{2.00}{1.63} & \ms{19.00}{1.47} & \ms{10.17}{0.85} \\

Seed-2.0-Pro & 
\ms{20.00}{2.83} & \ms{13.33}{3.77} & \ms{20.00}{2.83} & \ms{10.00}{1.63} & \ms{20.00}{5.89} & \ms{13.33}{4.99} & \ms{12.67}{0.94} & \ms{3.33}{0.94} & \ms{18.17}{2.87} & \ms{10.00}{1.87} \\

OpenAI o3 & 
\ms{24.00}{4.90} & \ms{18.67}{2.49} & \ms{9.33}{1.89} & \ms{4.67}{1.89} & \ms{10.67}{2.49} & \ms{7.33}{1.89} & \ms{2.00}{1.63} & \ms{1.33}{0.94} & \ms{11.50}{0.82} & \ms{8.00}{1.08} \\

Qwen3.5-Plus & 
\ms{20.67}{2.49} & \ms{4.00}{0.00} & \ms{20.00}{1.63} & \ms{6.67}{2.49} & \ms{28.67}{2.49} & \ms{14.67}{0.94} & \ms{16.00}{1.63} & \ms{2.67}{2.49} & \ms{21.33}{0.85} & \ms{7.00}{1.41} \\

DeepSeek-V3.2 & 
\ms{20.00}{3.27} & \ms{6.67}{0.94} & \ms{17.33}{0.94} & \ms{6.00}{0.00} & \ms{20.67}{7.54} & \ms{12.00}{3.27} & \ms{10.00}{1.63} & \ms{1.33}{0.94} & \ms{17.00}{2.55} & \ms{6.50}{0.82} \\

\bottomrule
\end{tabular}
}
\caption{Performance in thinking and non-thinking modes, ranked by overall PSR score. Standard deviations are shown in smaller font at the lower right of each mean. The best result in each column is highlighted in \textbf{bold}.}
\label{tab:performance}
% \vspace{-4mm}
\end{table*}

\paragraph{Substantial Length}
Given the substantial performance variation across context lengths~\citep{length}, we analyze the dataset's length distribution by tokenizing each instance (including tool descriptions) with the Qwen3 tokenizer. As shown in Figure~\ref{fig:length}, most instances fall within 3,000 to 7,000 tokens, averaging 4,754. Since models further interact through multiple turns, the effective context length continues to grow, posing a considerable challenge.

\paragraph{Precise Evaluation}
Since all tools are locally executable with annotated expected responses, we can precisely determine whether subqueries are resolved. Combined with our executable constraint validation module for code-based compliance checks at each step, this enables precise evaluation of problem-solving progress, constraint adherence, and self-refinement capability.

\section{Experimental Setup}
\label{sec:setup}

% In this section, we present the experimental setup used to evaluate LLMs with CCTU. Section~\ref{sec:models} describes the models under evaluation, Section~\ref{sec:metrics} defines the evaluation metrics, and Section~\ref{sec:details} provides the implementation details for the evaluation process.

\subsection{Models}
\label{sec:models}

We select nine top-tier models for evaluation: Claude Opus 4.6~\citep{Claude4.6}, DeepSeek-V3.2~\citep{deepseekv3.2}, Gemini 3 Pro~\citep{Gemini3}, GPT-5.1~\citep{GPT5.1}, GPT-5.2~\citep{GPT5.2}, Kimi 2.5~\citep{KimiK2.5}, OpenAI o3~\citep{OpenAIo3}, Qwen3.5-Plus~\citep{Qwen3.5}, and Seed-2.0-Pro~\citep{Seed2.0}. Each model is evaluated in both thinking and non-thinking modes.

\subsection{Metrics}
\label{sec:metrics}

Suppose there are $N$ test cases. The $i$-th test case contains $Q_i$ subqueries, where 
$q_{i,j}$ $\in$ $\{\texttt{unsolved},$ $\texttt{solved}\}$ indicates whether the $j$-th subquery is solved. 
Each test case is associated with $C_i$ constraints, and $c_{i,k}$ $\in$ $\{\texttt{unsat-}$ $\texttt{isfied},$ $\texttt{soft\texttt{-}satisfied},$ $\texttt{satisfied}\}$ denotes the status of the $k$-th constraint. 
Here, $\texttt{soft\texttt{-}satisfied}$ indicates that the constraint is initially violated but subsequently refined.
We evaluate model performance using two metrics.

\paragraph{Solve Rate (SR)}
Following~\citet{FTRL}, SR measures whether a model successfully solves all subqueries while satisfying all constraints.
\begin{multline*}
\mathrm{SR} =
\frac{1}{N}
\sum_{i=1}^{N}
\mathbb{I}\Big[
\bigg(\bigwedge_{j=1}^{Q_i} q_{i,j}=\texttt{solved}\bigg) \wedge
\bigg(\bigwedge_{k=1}^{C_i} \\
c_{i,k} \in \{\texttt{soft-satisfied}, \texttt{satisfied}\}
\bigg)
\Big]
\end{multline*}

\paragraph{Perfect Solve Rate (PSR)}
We introduce PSR as a stricter metric to measure whether a model solves all subqueries without any constraint violations.
\begin{multline*}
\mathrm{PSR} =
\frac{1}{N}
\sum_{i=1}^{N}
\mathbb{I}\Big[
\bigg(\bigwedge_{j=1}^{Q_i} q_{i,j}=\texttt{solved}\bigg) \wedge \\
\bigg(\bigwedge_{k=1}^{C_i} c_{i,k} = \texttt{satisfied}\bigg)
\Big]
\end{multline*}

\subsection{Implementation Details}
\label{sec:details}

To faithfully reflect each model's inherent capabilities, all models are invoked via their official API interfaces. Except for toggling between thinking and non-thinking modes, all inference hyperparameters are kept at their default values. To mitigate sampling-induced variability, we conduct three independent runs and report the mean and standard deviation of the results.

\begin{figure*}[!t]
    \centering
    \begin{subfigure}[!t]{0.45\linewidth}
        \centering
        \includegraphics[width=\linewidth]{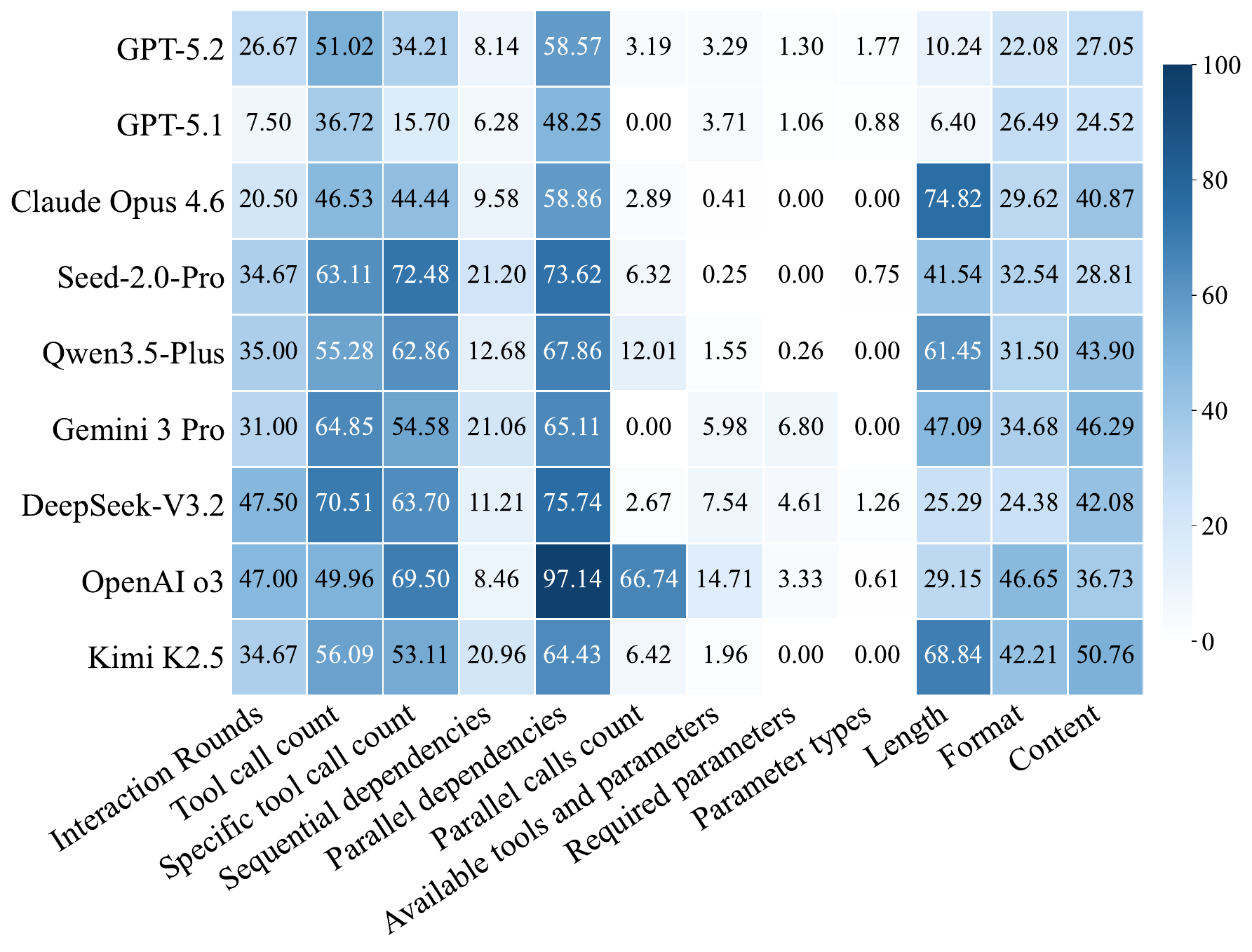}
        \caption{Thinking Mode}
    \end{subfigure}
    \quad
    \begin{subfigure}[!t]{0.45\linewidth}
        \centering
        \includegraphics[width=\linewidth]{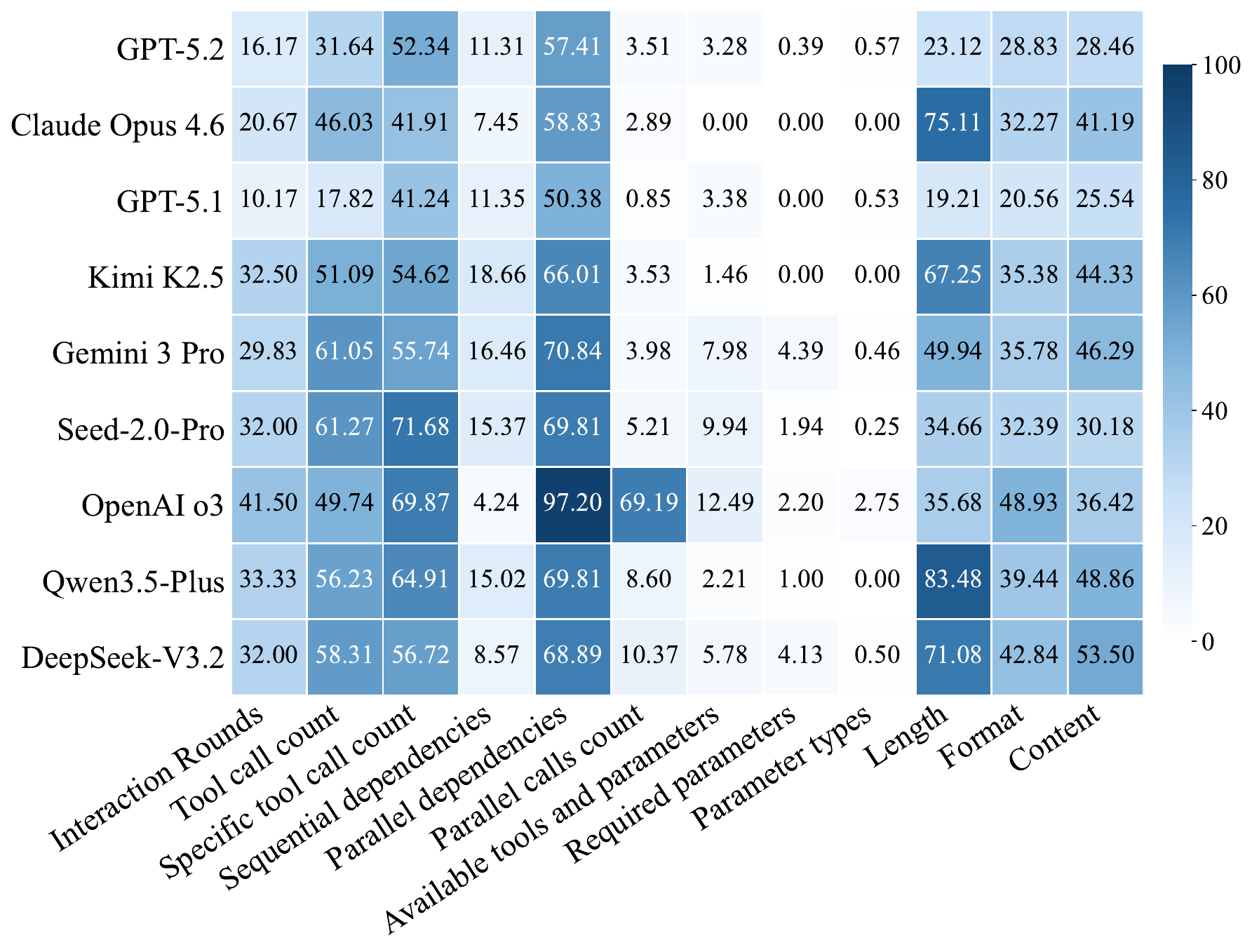}
        \caption{Non-Thinking Mode}
    \end{subfigure}
    \caption{The probability of different LLMs violating each category of constraints.}
    \label{fig:error}
% \vspace{-4mm}
\end{figure*}

\section{Experiments}
\label{sec:experiments}
% In this section, we analyze the performance of different LLMs on CCTU. Section~\ref{sec:main} presents the overall performance of the evaluated models. Section~\ref{sec:error} analyzes the violation rate under different constraint types. Section~\ref{sec:refine} further investigates the self-refinement capabilities of LLMs.

\subsection{Main Results}
\label{sec:main}

Table~\ref{tab:performance} presents the performance of various LLMs, from which we draw the following observations.

\textbf{CCTU presents substantial challenges for current LLMs while also clearly differentiating their capabilities.} Overall, the PSR of all LLMs remains below 20\%, with most models falling below 15\%. This indicates that current LLMs struggle to effectively use tools under constrained conditions. Moreover, performance in more complex scenarios is substantially worse than in simpler ones. For instance, GPT-5.2 achieves a PSR that is 14.67\% lower in parallel multi-hop tasks than in single-hop tasks. Although all LLMs perform poorly on CCTU, clear differences still emerge across models. In the thinking mode, the PSR of GPT-5.2 exceeds that of Kimi K2.5 by more than 10\%, highlighting notable disparities in the overall capabilities of different models.

\textbf{Although most models demonstrate improved performance in thinking mode, several exceptions are observed.} As expected, most models achieve higher performance in the thinking mode. For instance, Seed-2.0-Pro attains a PSR that is 4.83\% higher and an SR that is 2.16\% higher in thinking mode than in the non-thinking mode. However, notable exceptions are observed for Claude Opus 4.6 and Kimi K2.5. Through careful analysis, we find that this anomaly stems from the models' tendency to overthink in the thinking mode. Specifically, they sometimes rewrite parameter values specified in the prompt, and during the correction process they may also modify parameters that are originally correct, causing errors to persist. This issue occurs less frequently in the non-thinking mode.\footnote{Cases are provided in Appendix~\ref{sec:case}.}

\textbf{Models with high SR do not necessarily achieve high PSR, revealing two distinct performance strategies.} Although Claude Opus 4.6 achieves the highest SR, GPT-5.2 attains the highest PSR. Analysis reveals that these models adopt fundamentally different strategies. GPT-5.2 demonstrates stronger instruction-following capabilities, violating fewer constraints during execution and thus achieving higher PSR. However, when it does encounter errors, it struggles to self-refine, which limits its overall SR. In contrast, Claude Opus 4.6 commits more constraint violations during task execution but compensates through strong self-refinement ability, enabling it to correct its behavior based on feedback and ultimately complete more tasks. This contrast suggests that both robust instruction-following and effective self-refinement are essential.

\begin{figure*}[!t]
    \centering
    \begin{subfigure}[!t]{0.45\linewidth}
        \centering
        \includegraphics[width=\linewidth]{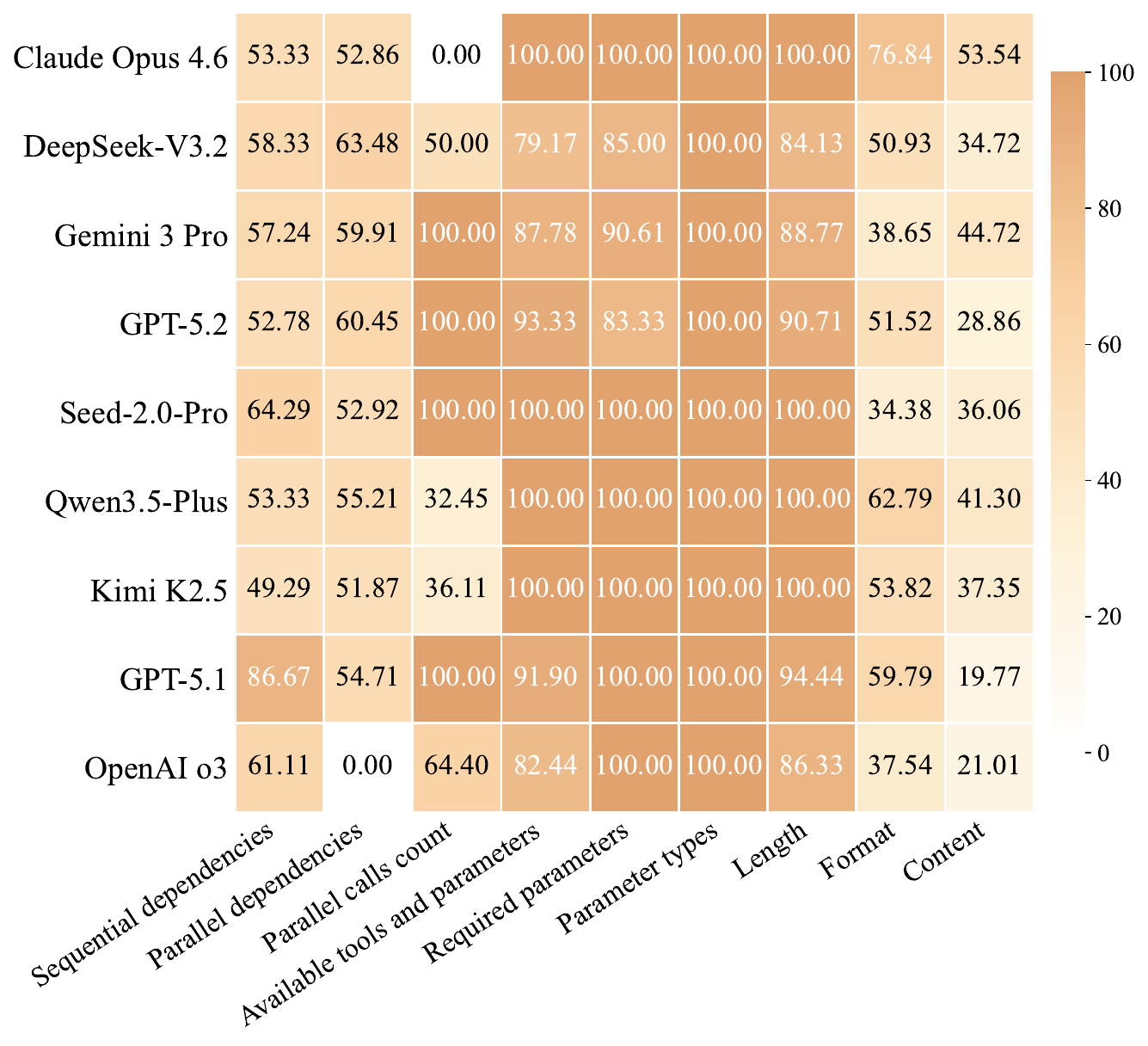}
        \caption{Thinking Mode}
    \end{subfigure}
    \quad
    \begin{subfigure}[!t]{0.45\linewidth}
        \centering
        \includegraphics[width=\linewidth]{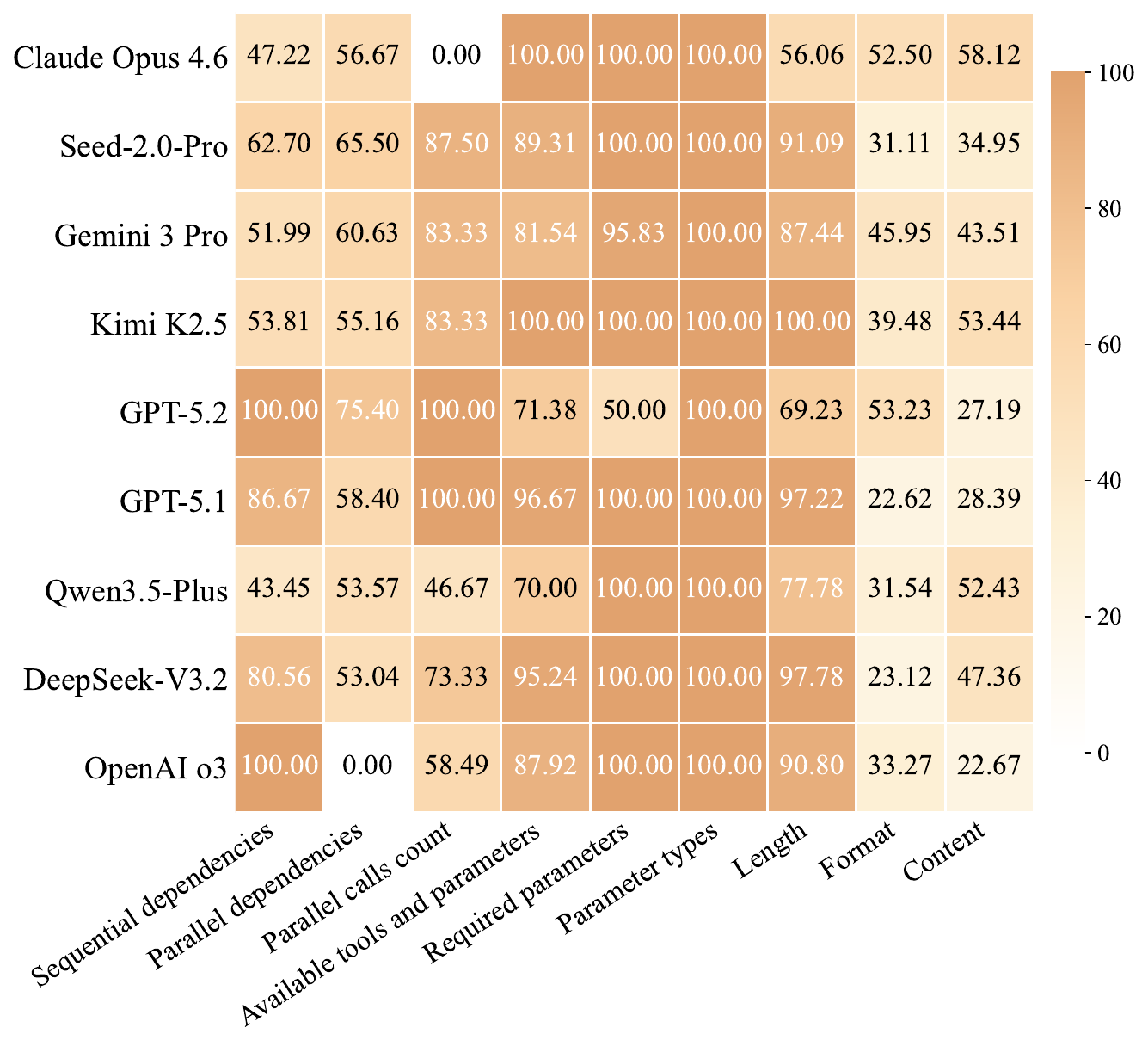}
        \caption{Non-Thinking Mode}
    \end{subfigure}
    \caption{Self-refinement probability of different LLMs under different constraint categories.}
    \label{fig:refine}
% \vspace{-4mm}
\end{figure*}

\subsection{Constraint Violation Analysis}
\label{sec:error}

Figure~\ref{fig:error} presents constraint violation rates across different categories, revealing several patterns.

\textbf{LLMs frequently violate constraints during tool use, particularly those in the resource and response dimensions.} Overall, all models exhibit constraint violations in more than 50\% of instances, with DeepSeek-V3.2 reaching as high as 86.83\%, indicating substantial room for improvement in constraint adherence during tool use. Across dimensions, models show notably higher violation rates in the resource and response dimensions. In the resource dimension, errors most commonly occur in tool call count constraints. We hypothesize that this stems from training paradigms that reward eventual task completion through trial-and-error, leading models to overlook restrictions on the number of tool invocations. In the response dimension, models primarily make mistakes related to response content. This is largely because models tend to summarize tool outputs in their final responses while failing to preserve key elements explicitly required in the original query.

\textbf{Error patterns shift unpredictably between thinking and non-thinking modes.} The thinking mode reduces a model's violation rate for some constraint types, while increases others. For instance, GPT-5.1 shows a lower violation rate for specific tool call count constraints in the thinking mode but a significantly higher rate for overall tool call count constraints. One explanation is that the extended reasoning enables more careful planning for specific tool allocation, yet the additional deliberation leads to more exploratory tool calls, exceeding the total invocation limit. More broadly, while thinking mode alters the distribution of error types, these shifts do not consistently translate into overall performance gains. These findings highlight the need for more effective thinking-mode mechanisms that can simultaneously reduce violations across all dimensions.

\textbf{Although less frequent, hallucinations and missing required parameters remain non-negligible issues.} Despite rapid advancements in current LLMs, most models struggle to fully understand the available tools and their parameters, leading to hallucination-related errors. In addition, models sometimes omit required parameters or provide incorrect parameter types during tool calls. Although such errors in the toolset dimension occur less frequently than those in other dimensions, they directly cause tool invocation failures and therefore remain critical issues that should not be overlooked.

\subsection{LLM Self-Refinement Analysis}
\label{sec:refine}

Figure~\ref{fig:refine} illustrates the proportion of errors that LLMs successfully correct through self-refinement, from which we derive the following findings.

\textbf{Even with detailed feedback guidance, models exhibit limited self-refinement capability under certain constraints.} The constraint validation module provides detailed feedback when models violate constraints. Overall, Claude Opus 4.6 achieves the highest correction rate at 65.36\%, while the correction rates of other models remain below 60\%, with OpenAI o3 reaching only 18.57\%. Although correction rates in the Toolset dimension are relatively high, some models still fail to reach 100\%, indicating limitations in their fundamental function-calling capabilities. Moreover, substantial room for improvement remains in both the behavior and response dimensions.

\textbf{Thinking modes enable some models to demonstrate stronger self-refinement capabilities, while degrading the performance of others.} Claude Opus 4.6, DeepSeek-V3.2, GPT-5.2, and Qwen3.5-Plus exhibit noticeably higher correction rates in the thinking mode than in the non-thinking mode. However, other models show no such improvement, and some even display lower correction rates when thinking is enabled. We attribute this to certain models doubling down on incorrect approaches during extended reasoning, which hinders subsequent correction.\footnote{Cases are provided in Appendix~\ref{sec:case}.} This observation further highlights the importance of well-designed thinking mechanisms.

\textbf{Effective self-refinement alone does not guarantee superior overall performance.} DeepSeek-V3.2 achieves an overall correction rate of 52.77\% in the thinking mode; however, Table~\ref{tab:performance} shows that its overall performance remains relatively low. This is primarily due to its higher probability of violating constraints and its comparatively weaker function-calling capabilities. In contrast, Claude Opus 4.6 benefits from strong function-calling capabilities and a significantly stronger self-refinement ability than other models, resulting in a higher SR. These results indicate that tool use under complex constraints is a challenging task that requires strong overall model capabilities.

\section{Conclusion}
\label{sec:conclusion}

In this paper, we introduce CCTU, a benchmark for tool use under complex constraints. CCTU is grounded in a four-dimensional, twelve-category constraint taxonomy. The benchmark comprises 200 carefully curated test cases, and is supported by an executable constraint validation module that enables step-level compliance checking during multi-turn interactions. We evaluate nine state-of-the-art LLMs in both thinking and non-thinking modes. Our results highlight the challenges CCTU poses for current models and reveal notable differences in instruction-following and self-refinement capabilities, providing valuable insights for advancing LLM development.

\section*{Limitations}

Although we have carefully designed a benchmark for tool use under complex constraints and conducted detailed evaluations of nine representative LLMs, our work still has several limitations.

\textbf{Despite our comprehensive constraint taxonomy, it does not cover all constraint categories encountered in real-world production environments.} The taxonomy introduced in Section~\ref{sec:taxonomy} includes 12 constraints across four dimensions, but it cannot capture every possible constraint. Nevertheless, these constraints represent common yet under-explored scenarios, and the insights derived from them remain valuable. In future work, the taxonomy can be further expanded by incorporating additional constraint types and constructing more challenging data, allowing the dataset to continue evolving and improving.

\textbf{The benchmark is constructed based on a single data source and therefore may not cover all possible tool use scenarios.} As described in Section~\ref{sec:construction}, CCTU is built upon FTRL. Therefore, its data distribution is limited by the characteristics of the source dataset. However, FTRL covers all combinations of subqueries, and our analysis shows that CCTU spans diverse domains, which partially mitigates this limitation.

\textbf{Constrained by the original dataset, CCTU currently contains only 200 test cases.} Since FTRL itself provides 200 instances, CCTU inherits the same scale. Nonetheless, these instances are carefully constructed, with each involving an average of seven constraint types and prompt lengths exceeding 4,700 tokens, resulting in a total evaluation surface substantially larger than the instance count alone suggests. Moreover, we propose an automated data generation pipeline that can be applied to other datasets in the future, enabling further expansion of the benchmark. To improve the reliability of our evaluation results, we also conduct three independent repeated experiments.

\section*{Acknowledgments}
The authors wish to thank the anonymous reviewers for their helpful comments.

% Bibliography entries for the entire Anthology, followed by custom entries
%\bibliography{custom,anthology-overleaf-1,anthology-overleaf-2}

% Custom bibliography entries only
\bibliography{custom}

\appendix

\section{Example of Constructed Data}
\label{sec:example}

As described in Section~\ref{sec:construction}, each sample in CCTU is carefully constructed and consists of the following core components:
\begin{itemize}
    \item \textbf{System Prompt:} Defines the agent’s role, operational protocol, and the complex constraints imposed on the task.
    \item \textbf{User Query:} The specific query that the agent is required to resolve.
    \item \textbf{Tools:} A JSON schema describing the available tools and their parameters.
    \item \textbf{Code Implementations:} Executable Python implementations of the provided tools that generate feedback when function calls are triggered.
    \item \textbf{Constraint Validation Codes:} Executable Python handlers used by the verification module to perform step-level constraint compliance checks and provide corrective feedback.
    \item \textbf{Data Source:} The structured scenario category from which the task is derived.
    \item \textbf{Answer:} The expected final result or options for resolving the user query.
    \item \textbf{Unresolved Set:} The prerequisite tools required to resolve the query and their expected execution outputs.
\end{itemize}

To help readers better understand the structure and content of the benchmark, we provide a complete and representative test case below in Figure~\ref{fig:case}.

\section{Pipeline for Benchmark Construction}
\label{sec:pipeline}

As described in Section~\ref{sec:construction}, we design a comprehensive data construction pipeline to build high-quality evaluation data. To illustrate this pipeline more intuitively, we summarize it in Algorithm~\ref{alg:construct}.

\newpage
\onecolumn
% \vspace{0.5em}
\begin{center}
% --- 1. System Prompt ---
\begin{tcolorbox}[colback=teal!5, colframe=teal!50, title=\textbf{System Prompt}, arc=1mm, boxrule=0.5mm, breakable]
\small
\textit{*Note: The text highlighted in \textcolor{red}{\textbf{red}} represents the injected complex constraints across different dimensions.}
% \vspace{1em}

{\large \textbf{Role and Operational Context}}
\vspace{0.2em}\hrule% \vspace{0.5em}
You are an advanced Autonomous Historical Temporal Analysis Agent. You act as an independent problem-solving engine designed to resolve queries about the chronological relationships between cultural monuments, technological innovations, and philosophical movements. Your primary function is to use external tools to verify historical timelines and establish causal interdependencies between societal developments and technological breakthroughs.

% \vspace{1em}
{\large \textbf{Core Philosophy and Knowledge Constraints}}
\vspace{0.2em}\hrule% \vspace{0.5em}
You operate under a strict tool-dependency protocol. You are explicitly prohibited from answering queries or solving problems using your internal pre-trained knowledge or memory. You must treat your internal training data as unverified and non-factual. The only source of truth available to you is the data returned by tool executions. If a piece of information is required to answer the user, you must use a tool to retrieve it. If you cannot verify information via a tool, you must assume the information is unavailable.

% \vspace{1em}
{\large \textbf{Autonomous Workflow Architecture}}
\vspace{0.2em}\hrule% \vspace{0.5em}
Upon receiving the user's initial query, you must enter a continuous, self-regulated execution loop consisting of task decomposition, action, observation, and reflection. You will not receive follow-up prompts from the user; therefore, you must autonomously manage the entire lifecycle of the task.

% \vspace{0.5em}
\textbf{1. Complex Task Decomposition}\\
When specific objectives are complex, you must break them down into smaller, atomic sub-tasks. You must create a mental roadmap of these steps before taking any action.

% \vspace{0.5em}
\textbf{2. Dependency Analysis and Parallel Execution}\\
To maximize efficiency, you are required to analyze the dependencies between your planned sub-tasks.
\begin{itemize}
    \setlength{\itemsep}{0pt}
    \setlength{\parskip}{0pt}
    \setlength{\parsep}{0pt}
    \item \textbf{Dependent Tasks:} If Task B requires the output of Task A, you must execute them sequentially.
    \item \textbf{Independent Tasks:} If multiple tasks do not rely on each other's outputs, you are expected to execute them simultaneously in a single interaction turn. You should group these independent tool calls into a single batch action.
\end{itemize}

% \vspace{0.5em}
\textbf{3. Precision in Tool Selection}\\
You must rigorously evaluate the available tools to select the most appropriate function for each sub-task. You are responsible for ensuring that all parameter inputs adhere strictly to the required schema (types, formats, and constraints).

% \vspace{0.5em}
\textbf{4. Interaction Round Limit and Forced Termination}\\
\textcolor{red}{\textbf{You operate within a strict budget of maximum 20 interaction rounds (indexed starting from Round 1).}} A ``Round'' is defined as a single cycle of processing an input (the User Question or previous Tool Observations) and producing a corresponding output. This output can be either a tool execution request ([ACTION]) OR a final response ([FINAL ANSWER]).\\
\textbf{CRITICAL STOP PROTOCOL:} If you reach upper limit round, you must immediately cease all further tool executions, regardless of whether the task is complete or if further tools are needed. You are strictly prohibited from proceeding to a more round. You must accept the current state and provide the best possible [FINAL ANSWER] based solely on the information accumulated up to that point.

% \vspace{1em}
{\large \textbf{Error Resolution and Persistence Strategy}}
\vspace{0.2em}\hrule% \vspace{0.5em}
You must adopt a resilient approach to failure. If a tool execution fails or returns an error, you must not abandon the task. Instead, you must engage in an iterative retry process:
\begin{itemize}
    \setlength{\itemsep}{0pt}
    \setlength{\parskip}{0pt}
    \setlength{\parsep}{0pt}
    \item Analyze the specific error message to understand the root cause (e.g., invalid parameter, timeout, incorrect format).
    \item Formulate a corrected plan, such as modifying the parameters, trying a different search term, or selecting an alternative tool that achieves the same outcome.
    \item Continue this cycle of reflection and retry until the sub-task is successfully completed.
\end{itemize}

% \vspace{1em}
{\large \textbf{Execution Protocol and Process Constraints}}
\vspace{0.2em}\hrule% \vspace{0.5em}
You must strictly adhere to the following multi-stage execution protocol for every interaction turn. These are mandatory behavioral constraints that define your operational lifecycle.

% \vspace{0.5em}
\textbf{[THOUGHT]}\\
This phase serves as your strategic foundation, designed to ensure every action is calculated and effective. You must use this space to decompose the complex objective into manageable sub-tasks, creating a clear mental roadmap. Focus on analyzing the logical flow to identify which steps require sequencing and which offer opportunities for parallel execution to optimize efficiency. By articulating your rationale for tool and parameter choices here, you establish a robust plan that minimizes errors during the execution phase. \textcolor{red}{\textbf{You must ensure the \texttt{philosopher\_concept\_identifier} tool is invoked at most once during the process.}}

% \vspace{0.5em}
\textbf{[ACTION]}\\
This section is strictly designated for invoking external tools based on your preceding plan. If your dependency analysis identified independent tasks, you are required to execute them in parallel within this single turn to maximize operational efficiency. Every tool call must strictly adhere to the definitions provided in available tools or the native function calling convention, as invalid parameters or schema violations constitute a critical protocol failure. \textcolor{red}{\textbf{The \texttt{philosopher\_concept\_identifier} tool must be invoked before the \texttt{historical\_figure\_info} tool.}}

% \vspace{0.5em}
\textbf{[REFLECTION]}\\
Upon receiving tool observations, you must engage in critical analysis to validate the data before proceeding. You must determine if the returned information is sufficient to resolve the user's request or if specific errors require a remediation plan, such as retrying with adjusted parameters. This phase dictates whether the workflow loops back to the planning phase for further steps or proceeds to a final resolution.

% \vspace{0.5em}
\textbf{[FINAL ANSWER]}\\
You are authorized to output this section only when the user's objective is fully satisfied by the collected data. The response must be derived strictly from the tool observations without adding external information or hallucinations. Provide the specific answer or data requested by the user directly, avoiding unnecessary preamble or summary. \textcolor{red}{\textbf{The final answer must end with a period (.) to ensure proper sentence closure.}}
\end{tcolorbox}

% \vspace{1em}

% --- 2. User Question ---
\begin{tcolorbox}[colback=gray!5, colframe=gray!40, title=\textbf{User Query}, arc=1mm, boxrule=0.5mm, breakable]
\small
Please call given tools to answer the question. Please note that all your information must be obtained by calling tools and not by answering the question directly. If the call fails, you need to try to correct it and continue until you arrive at an answer.\\
\\
\textbf{Question:} Which is older: (a) the monument built in the province governed by the politician who initiated reforms inspired by the movement originating from the region where the philosopher famous for his allegory was born, or (b) the invention of the telephone?
\end{tcolorbox}

% \vspace{1em}

% --- 3. Tools ---
\begin{tcolorbox}[colback=blue!5, colframe=blue!40, title=\textbf{Tools}, arc=1mm, boxrule=0.5mm, breakable]
\scriptsize
\begin{verbatim}
[
  {
    "type": "function",
    "function": {
      "name": "philosopher_concept_identifier",
      "description": "A sophisticated tool designed to identify philosophers 
                      based on their notable concepts...",
      "parameters": {
        "type": "object",
        "properties": {
          "concept": { "type": "string", "description": "..." },
          "era": { "type": "string", "enum": ["Ancient", "Medieval", ...] }
        },
        "required": ["concept"]
      }
    }
  },
  {
    "type": "function",
    "function": {
      "name": "historical_figure_info",
      "description": "A tool designed to retrieve detailed biographical 
                      information about historical figures...",
      "parameters": {
        "type": "object",
        "properties": {
          "figure_name": { "type": "string", "description": "..." },
          "info_type": { "type": "string", "enum": ["birthplace", ...] }
        },
        "required": ["figure_name"]
      }
    }
  }
  
  // [... JSON schemas for 11 additional tools (e.g., historical_information_retriever, 
  // monument_locator) omitted for brevity ...]
]
\end{verbatim}
\end{tcolorbox}

% \vspace{1em}

% --- 4. Tool Codes ---
\begin{tcolorbox}[colback=purple!5, colframe=purple!40, title=\textbf{Code Implementations}, arc=1mm, boxrule=0.5mm, breakable]
\scriptsize
\begin{verbatim}
def philosopher_concept_identifier(concept, era=None, region=None, 
                                   include_minor_works=False, work_type=None, 
                                   philosophical_school=None, influence_level=None):
    """
    Identifies philosophers based on their notable concepts, theories, or works.
    """
    # Error handling for required parameter
    if not concept:
        return "Error: 'concept' is a required parameter."

    # [... Error handling for enumerated parameters omitted for brevity ...]

    # Simulated logic for identifying philosophers
    if concept.lower() == 'allegory':
        return "Plato is famous for his allegory, particularly the Allegory of the Cave."
    
    # Default response if no specific logic matches
    return "No specific philosopher identified for the given concept."

# [... Executable Python codes for 11 additional tools (e.g., historical_figure_info, 
# historical_governance_finder) omitted for brevity ...]
\end{verbatim}
\end{tcolorbox}

% \vspace{1em}

% --- 5. Validation Code (New Section) ---
\begin{tcolorbox}[colback=cyan!5, colframe=cyan!50, title=\textbf{Constraint Validation Codes}, arc=1mm, boxrule=0.5mm, breakable]
\scriptsize
\begin{verbatim}
class MaxCallsPerToolHandler(BaseHandler):
    key = ("tool", "max calls per tool")

    def configure(self, checker: Any, idx: int) -> None:
        data = _load_json(checker._json_check_file(idx))
        for tool_name, call_times in data["max_calls_per_tool"].items():
            if tool_name not in checker.tool_name_list:
                raise KeyError(f"tool {tool_name} not in tool list {checker.tool_name_list}")
            checker.max_callTimesPerTool[tool_name] = to_int(call_times)

    def check(self, checker: Any, ctx: TurnContext, fb: Any) -> None:
        if ctx.is_final:
            return
        for call in ctx.tool_calls or []:
            name = call["function"]["name"]
            checker.callTimesPerTool[name] += 1
            if checker.callTimesPerTool[name] > checker.max_callTimesPerTool[name]:
                fb.add_tool(
                    call.get("id", ""),
                    f"INSTRUCTION FOLLOWING ERROR: MAX CALLS PER TOOL NOT FOLLOWED! "
                    f"Maximum call tool '{name}' times requirement not met: called "
                    f"{checker.callTimesPerTool[name]} times, requires at most "
                    f"{checker.max_callTimesPerTool[name]}."
                )

# [... Python constraint handlers for 11 additional constraint categories (e.g., 
# ToolOrderHandler, ToolParallelHandler) omitted for brevity ...]
\end{verbatim}
\end{tcolorbox}

% \vspace{1em}

% --- 6. Ground Truth & Metadata ---
\begin{tcolorbox}[colback=gray!5, colframe=gray!40, title=\textbf{Others}, arc=1mm, boxrule=0.5mm, breakable]
\small
\textbf{Data Source (Scenario Category):} Parallel Multi-Hop
% % \vspace{0.5em}

\textbf{Answer:} a

% % \vspace{0.5em}
\textbf{Unsolved Set (Required Execution Trajectory):}
\scriptsize
\begin{verbatim}
{
  "philosopher_concept_identifier": ["Plato"],
  "historical_figure_info": ["Athens"],
  "movement_origin_identifier": ["Democracy"],
  "political_reform_initiator_finder": ["Benjamin Franklin"],
  "historical_governance_finder": ["Pennsylvania"],
  "monument_locator": ["Liberty Bell"],
  "historical_information_retriever": ["1876", "1752"]
}
\end{verbatim}
\end{tcolorbox}
\captionof{figure}{A test case of CCTU.}
\label{fig:case}
\end{center}

\newpage

\begin{algorithm}[!t]
\caption{Pipeline for Benchmark Construction}
\label{alg:construct}
\begin{algorithmic}[1]
\Require Initial dataset $\mathcal{D}$, constraint
  taxonomy $\mathcal{C}$, model $M$
\Ensure Constrained dataset $\mathcal{D'}$ with validation code

\For{each instance $d \in \mathcal{D}$}
    \State Extract unsolved subqueries $\mathcal{Q}$, toolset
      $\mathcal{T}$, scenario category $s$
      \Comment{Parse original instance}
    \Repeat
    \Comment{Reference trajectory generation}
        \State Sample trajectory $\tau$ using
          $M(d)$
        \State Update $\mathcal{Q}$ according to solved
          subqueries
    \Until{$\mathcal{Q} = \emptyset$}
    \State Store $\tau$
    
    \vspace{0.3em}
    \State $\mathcal{K} \gets \emptyset$
      \Comment{Initialize constraint set}
    \For{each constraint type $c \in \mathcal{C}$
      except Toolset}
    \Comment{Controlled constraint expansion}
        \If{Bernoulli$(0.5)$ and compatible with
          scenario $s$}
            \State Inject constraint $c$ using $M$
              conditioned on $\tau$
            \State $\mathcal{K} \gets \mathcal{K}
              \cup \{c\}$
        \EndIf
    \EndFor

    \vspace{0.3em}
    \Repeat
    \Comment{LLM-based filtering}
        \State Verify consistency of $(d,\mathcal{K})$
          using $M$
        \If{conflict detected}
            \State Revise constraints
        \EndIf
    \Until{verification succeeds}

    \vspace{0.3em}
    \State Generate constraint-free task context $ctx$ using $M$
      \Comment{Task context integration}
    \State Construct constrained instance $d'=(ctx,d,\mathcal{K})$

    \vspace{0.3em}
    \For{each constraint $k \in \mathcal{K}$}
    \Comment{Constraint validation code generation}
        \State Generate validation code $v_k$ using $M$
    \EndFor
    \State Add $(d',\{v_k\})$ to $\mathcal{D'}$
\EndFor

\vspace{0.3em}
\For{each instance $(d',V) \in \mathcal{D'}$}
    \Repeat
    \Comment{Data verification}
        \State Annotator reviews $d'$
        \If{issues found}
            \State Revise $d'$
        \EndIf
        \State Next annotator reviews $d'$
    \Until{two consecutive annotators agree}

    \vspace{0.3em}
    \Repeat
    \Comment{Code verification}
        \State Annotator reviews $V$
        \If{issues found}
            \State Revise $V$
        \EndIf
        \State Next annotator reviews $V$
    \Until{two consecutive annotators agree}
\EndFor

\vspace{0.3em}
\State \Return $\mathcal{D'}$
\end{algorithmic}
\end{algorithm}

\newpage
\twocolumn

\section{Details of Human Annotation}
\label{sec:annotatation}

\subsection{Overview}

\begin{table}[h]
    \centering
    \resizebox{\linewidth}{!}
    {
    \begin{tabular}{lcccc}
        \toprule
        \textbf{Stage} & \textbf{One-Round} & \textbf{Two-Round} & \textbf{Total} & \textbf{Revision Rate} \\
        \midrule
        Data Verification & 38 & 6 & 44 & 21\% \\
        Code Verification & 19 & 35 & 54 & 27\% \\
        \bottomrule
    \end{tabular}
    }
    \caption{Statistics of manual revisions during the quality control process. The revision rate represents the proportion of revised instances relative to the total dataset (200 instances).}
    \label{tab:revision}
\end{table}

As described in Section~\ref{sec:construction}, we conduct iterative human annotation on the constructed data and code through a carefully designed process. The annotation continues until two consecutive annotators independently agree that the data has passed inspection, thereby ensuring its quality. During this process, six graduate students majoring in computer science are recruited to participate in the annotation task, with each annotator working independently. All annotations are completed within three weeks.

Table~\ref{tab:revision} presents the proportion of data that requires manual modification. The results show that more than 70\% of the data requires no manual intervention, and the number of modification rounds does not exceed two. This demonstrates both the precision and practicality of our data construction process. Furthermore, because we employ Qwen3-32B, which is not specifically designed for code generation, the proportion of code requiring modification is slightly higher than that of the instructions.

\subsection{Identified Issues during Data Verification}

During the data verification stage, the identified issues can be categorized into three primary types: conflicts among constraints, mismatches between constraints and scenario structures, and conflicts between constraints and tools. To illustrate these issues, we present representative examples of manual corrections for each category below.

% ---------------------------------------------------------
% Example 1: Conflicts Among Constraints
% ---------------------------------------------------------
\paragraph{Conflicts Among Constraints}
As shown in Figure~\ref{fig:conflict}, the system prompt generated by the LLM requires the model's response to be a valid JSON object while simultaneously mandating that the response end with a period, creating a conflict between the constraints. Through manual annotation, we revise one of these constraints to require that the response contain a JSON object, rather than strictly enforcing that the entire output conform to JSON format. This modification resolves the conflict while still enabling the evaluation of the model's ability to understand composite constraints.

% ---------------------------------------------------------
% Example 2: Mismatches Between Constraints and Scenario Structures
% ---------------------------------------------------------
\paragraph{Mismatches Between Constraints and Scenario Structures}
As shown in Figure~\ref{fig:mismatch}, this instance belongs to the single-hop scenario. However, because LLMs reference trajectories that include trial-and-error steps, a requirement of at least three rounds of interaction is initially imposed. To ensure that the task remains logically consistent and solvable within the single-hop structure, this lower bound is converted into an upper bound during human annotation.

% ---------------------------------------------------------
% Example 3: Conflicts Between Constraints and Tool Requirements
% ---------------------------------------------------------
\paragraph{Conflicts Between Constraints and Tools}
As shown in Figure~\ref{fig:tool}, resolving the user's query requires invoking the \textsf{product\_and\_appliance\_specification\_retriever} tool twice to retrieve specifications for two distinct devices. However, the constraint generated by the LLM limits the total number of tool invocations to at most once, which conflicts with the operational requirements. Therefore, after human annotation, the maximum permitted number of tool invocations is adjusted from one to two to ensure task solvability.

\subsection{Identified Issues during Code Verification}

During the code verification stage, the identified issues can be categorized into two primary types: overly strict evaluation criteria and redundant logic across different constraint handlers. To illustrate these issues, we present representative examples of manual corrections below.

% ---------------------------------------------------------
% Example 1: Overly Strict Evaluation Criteria
% ---------------------------------------------------------
\paragraph{Overly Strict Evaluation Criteria}
As shown in Figure~\ref{fig:strict}, the validation code generated by the LLM for the Markdown format constraint is excessively rigid. It requires the model's response to simultaneously feature a heading, a list, and text emphasis to pass the check. Through manual annotation, we relax this criterion so that the presence of at least one of these Markdown elements constitutes a success. This modification prevents false negatives and ensures a fairer evaluation of the model's formatting capabilities.

% ---------------------------------------------------------
% Example 2: Redundant Validation Logic
% ---------------------------------------------------------
\paragraph{Redundant Validation Logic}
As shown in Figure~\ref{fig:redundant}, the LLM occasionally mixes validation logic for multiple distinct constraints into a single function. Specifically, the validation code intended solely for the JSON format constraint incorrectly incorporates word-count checks corresponding to the response length constraint. To maintain modularity and prevent duplicate penalization during evaluation, human annotators remove the redundant length validation logic from the format checker.

\section{Case Studies}
\label{sec:case}

In Section~\ref{sec:experiments}, we find that although most LLMs exhibit performance improvements when switching from the non-thinking mode to the thinking mode, several models do not benefit from this change, particularly Claude Opus 4.6 and Kimi K2.5. Further analysis indicates that this behavior is largely attributable to overthinking. Figures~\ref{fig:claudeover} and~\ref{fig:kimiover} illustrate this phenomenon.

In Figure~\ref{fig:claudeover}, under the non-thinking mode, Claude Opus 4.6 successfully obtains the correct information by adjusting parameters after the first query fails. In contrast, under the thinking mode, the model modifies multiple parameters simultaneously in each attempt, leading to incorrect parameter combinations and ultimately failing to resolve the user's query. Figure~\ref{fig:kimiover} presents a similar case for Kimi K2.5. When querying the specific location of a clash, the model successfully obtains the answer in the non-thinking mode by sequentially testing the two countries—Israel and Palestine—mentioned in the query. However, in the thinking mode, the model overcomplicates the reasoning process and attempts to infer alternative locations rather than directly using those provided in the query. As a result, it fails to retrieve relevant information and is unable to complete the task.

In addition, the self-refinement capabilities of some models do not improve in the thinking mode. Our analysis suggests that during the extended reasoning, the models often become more confident in their initial incorrect solutions, which prevents them from adjusting their behavior based on feedback. As illustrated in Figure~\ref{fig:seedself}, although Seed-2.0-Pro receives feedback instructing it to perform parallel calls to specific tools, the model ignores this mandatory constraint during its reasoning process. Instead, it assumes that the tool has already been called earlier and therefore does not need to be invoked again, which ultimately prevents it from successfully retrieving the required information.

\section{Prompts for Benchmark Construction}
\label{sec:prompt}

As described in Section~\ref{sec:construction}, we utilize Qwen3-32B to assist in the data construction process. We present the prompt templates used in the pipeline from Figure~\ref{fig:promptconstraintexpansion} to Figure~\ref{fig:promptformatcontent}.

\newpage
\onecolumn

\begin{center}
\begin{tcolorbox}[colback=white, colframe=black!60, arc=2mm, boxrule=0.8mm, breakable]

    \begin{tcolorbox}[colback=gray!5, colframe=gray!40, title=\textbf{Task Information}, arc=1mm, boxrule=0.5mm]
    \textbf{User Question:} Which two countries signed an agreement to trade in their own currencies in March 2023?\\
    \textbf{Tools:} ...\\
    \textbf{Unsolved Set:} ...\\
    \textbf{Data Source:} ...
    \end{tcolorbox}
    
    % \vspace{0.5em}

    \begin{tcolorbox}[colback=red!5, colframe=red!40, title=\textbf{LLM-Generated System Prompt}, arc=1mm, boxrule=0.5mm]
    [... Background and other constraints omitted for brevity ...]
    \newline - The answer must be formatted as a valid JSON object containing ``countries'' (array of two country names) and ``date'' (string in ``March 2023'' format) fields.
    \newline - Contain between 15 and 20 words (inclusive), end with a period, and use a comma to separate the two country names.
    \newline [...]
    \end{tcolorbox}
    
    % \vspace{0.5em}

    \begin{tcolorbox}[colback=teal!5, colframe=teal!50, title=\textbf{Human-Calibrated System Prompt}, arc=1mm, boxrule=0.5mm]
    [... Background and other constraints omitted for brevity ...]
    \newline - The answer must include a valid JSON object containing ``countries'' (array of two country names) and ``date'' (string in ``March 2023'' format) fields.
    \newline - Contain between 15 and 20 words (inclusive), end with a period, and use a comma to separate the two country names.
    \newline [...]
    \end{tcolorbox}

\end{tcolorbox}
\captionof{figure}{An example illustrating a conflict among constraints.}
\label{fig:conflict}
\end{center}

\newpage

\begin{center}
\begin{tcolorbox}[colback=white, colframe=black!60, arc=2mm, boxrule=0.8mm, breakable]

    \begin{tcolorbox}[colback=gray!5, colframe=gray!40, title=\textbf{Task Information}, arc=1mm, boxrule=0.5mm]
    \textbf{User Question:} Which president won the 2023 Egyptian presidential election held between December 10-12, 2023?\\
    \textbf{Tools:} ...\\
    \textbf{Unsolved Set:} ...\\
    \textbf{Data Source:} Single-Hop
    \end{tcolorbox}
    
    % \vspace{0.5em}

    \begin{tcolorbox}[colback=red!5, colframe=red!40, title=\textbf{LLM-Generated System Prompt}, arc=1mm, boxrule=0.5mm]
    [... Background and other constraints omitted for brevity ...]
    \newline - You are required to engage in at least 3 interaction rounds (indexed starting from Round 1).
    \newline [...]
    \end{tcolorbox}
    
    % \vspace{0.5em}

    \begin{tcolorbox}[colback=teal!5, colframe=teal!50, title=\textbf{Human-Calibrated System Prompt}, arc=1mm, boxrule=0.5mm]
    [... Background and other constraints omitted for brevity ...]
    \newline - You operate within a strict budget of maximum 3 interaction rounds (indexed starting from Round 1).
    \newline [...]
    \end{tcolorbox}

\end{tcolorbox}
\captionof{figure}{An example illustrating a mismatch between the constraints and the scenario structure.}
\label{fig:mismatch}
\end{center}

\begin{center}
\begin{tcolorbox}[colback=white, colframe=black!60, arc=2mm, boxrule=0.8mm, breakable]

    \begin{tcolorbox}[colback=gray!5, colframe=gray!40, title=\textbf{Task Information}, arc=1mm, boxrule=0.5mm]
    \textbf{User Question:} What is the total wattage (W) of GE Profile PHP900 induction cooktop and Bosch 800 Series induction range?\\
    \textbf{Tools:} [\dots, \textsf{product\_and\_appliance\_specification\_retriever}, \dots]\\
    \textbf{Unsolved Set:} \texttt{\{"product\_and\_appliance\_specification\_retriever": ["9,700 W", "5,000 W"]\}}\\
    \textbf{Data Source:} Parallel Single-Hop
    \end{tcolorbox}
    
    % \vspace{0.5em}

    \begin{tcolorbox}[colback=red!5, colframe=red!40, title=\textbf{LLM-Generated System Prompt}, arc=1mm, boxrule=0.5mm]
    [... Background and other constraints omitted for brevity ...]
    \newline - You are strictly limited to a maximum of 1 tool call during the entire execution.
    \newline [...]
    \end{tcolorbox}
    
    % \vspace{0.5em}

    \begin{tcolorbox}[colback=teal!5, colframe=teal!50, title=\textbf{Human-Calibrated System Prompt}, arc=1mm, boxrule=0.5mm]
    [... Background and other constraints omitted for brevity ...]
    \newline - You are strictly limited to a maximum of 2 tool calls during the entire execution.
    \newline [...]
    \end{tcolorbox}

\end{tcolorbox}
\captionof{figure}{An example illustrating a conflict between constraints and tools.}
\label{fig:tool}
\end{center}

\begin{center}
\begin{tcolorbox}[colback=white, colframe=black!60, arc=2mm, boxrule=0.8mm, breakable]

    \begin{tcolorbox}[colback=gray!5, colframe=gray!40, title=\textbf{Task Information}, arc=1mm, boxrule=0.5mm, breakable]
    \textbf{Target Constraint:} Markdown Format\\
    \textbf{Constraint Definition:} The agent's entire response must be formatted using Markdown syntax, ensuring proper use of elements such as headings, lists, bold/italic text, links, and code blocks to enhance readability.
    \end{tcolorbox}
    
    % \vspace{0.5em}

    \begin{tcolorbox}[colback=red!5, colframe=red!40, title=\textbf{LLM-Generated Validation Code}, arc=1mm, boxrule=0.5mm, breakable]
\scriptsize
\begin{verbatim}
[... Helper utilities and regex patterns omitted for brevity ...]

def validate_format(response: str) -> Tuple[bool, str]:
    if not response or not response.strip():
        return False, "Response is empty."

    issues = []
    # Requires ALL elements to be present
    if not has_markdown_heading(response):
        issues.append("Missing a Markdown heading.")
    if not has_markdown_list(response):
        issues.append("Missing a Markdown list.")
    if not has_emphasis(response):
        issues.append("Missing emphasis.")
    if not has_balanced_emphasis_markers(response):
        issues.append("Unbalanced emphasis markers detected.")

    if issues:
        return False, "Markdown formatting requirements not met: " + " ".join(issues)

    return True, "Markdown format validated."
\end{verbatim}
    \end{tcolorbox}
    
    % \vspace{0.5em}

    \begin{tcolorbox}[colback=teal!5, colframe=teal!50, title=\textbf{Human-Calibrated Validation Code}, arc=1mm, boxrule=0.5mm, breakable]
\scriptsize
\begin{verbatim}
[... Helper utilities and regex patterns omitted for brevity ...]

def validate_format(response: str) -> Tuple[bool, str]:
    if not response or not response.strip():
        return False, "Response is empty."

    if not has_balanced_emphasis_markers(response):
        return False, "Unbalanced emphasis markers detected."

    # Relaxed to require AT LEAST ONE element
    has_heading = has_markdown_heading(response)
    has_list = has_markdown_list(response)
    has_emp = has_emphasis(response)
    
    has_any_formatting = has_heading or has_list or has_emp
    
    if not has_any_formatting:
        return False, "Missing Markdown formatting. Include at least one element."
    
    return True, "Markdown format validated: Found at least one formatting element."
\end{verbatim}
    \end{tcolorbox}

\end{tcolorbox}
\captionof{figure}{An example illustrating the correction of overly strict evaluation criteria.}
\label{fig:strict}
\end{center}

\newpage

\begin{center}
\begin{tcolorbox}[colback=white, colframe=black!60, arc=2mm, boxrule=0.8mm, breakable]

    \begin{tcolorbox}[colback=gray!5, colframe=gray!40, title=\textbf{Task Information}, arc=1mm, boxrule=0.5mm]
    \textbf{Target Constraint:} JSON Format\\
    \textbf{Constraint Definition:} The response must be structured as a valid JSON object, adhering to proper syntax rules.
    \end{tcolorbox}
    
    % \vspace{0.5em}

    \begin{tcolorbox}[colback=red!5, colframe=red!40, title=\textbf{LLM-Generated Validation Code}, arc=1mm, boxrule=0.5mm, breakable]
\scriptsize
\begin{verbatim}
[... JSON parsing utilities omitted for brevity ...]

def validate_format(response: str) -> Tuple[bool, str]:
    trimmed = response.strip()
    if not (trimmed.startswith("{") and trimmed.endswith("}")):
        return False, "The output must be a single JSON object."

    try:
        obj = _parse_json_object(response)
    except ValueError as e:
        return False, str(e)
        
    # Redundant logic: Validating length constraint inside format checker
    word_count = _count_words_in_strings(_iter_string_values(obj))
    if word_count < 10:
        return False, f"Word-count constraint violated: found {word_count} words."
    if word_count > 30:
        return False, f"Word-count constraint violated: found {word_count} words."

    return True, "OK: Response is a valid JSON object with 10–30 words."
\end{verbatim}
    \end{tcolorbox}
    
    % \vspace{0.5em}

    \begin{tcolorbox}[colback=teal!5, colframe=teal!50, title=\textbf{Human-Calibrated Validation Code}, arc=1mm, boxrule=0.5mm, breakable]
\scriptsize
\begin{verbatim}
[... JSON parsing utilities omitted for brevity ...]

def validate_format(response: str) -> Tuple[bool, str]:
    trimmed = response.strip()
    if not (trimmed.startswith("{") and trimmed.endswith("}")):
        return False, "The output must be a single JSON object."

    # Strict parsing with duplicate key detection and structure checks.
    try:
        obj = _parse_json_object(response)
    except ValueError as e:
        return False, str(e)

    # Redundant length validation logic has been successfully removed.

    return True, "OK: Response is a valid JSON object with unique keys."
\end{verbatim}
    \end{tcolorbox}

\end{tcolorbox}
\captionof{figure}{An example illustrating the removal of redundant validation logic.}
\label{fig:redundant}
\end{center}

\newpage

\begin{center}
\begin{tcolorbox}[colback=white, colframe=black!70, arc=2mm, boxrule=0.8mm, breakable]

    % --- Task Context ---
    \begin{tcolorbox}[colback=gray!5, colframe=gray!40, title=\textbf{Task Information}, arc=1mm, boxrule=0.5mm]
    \small
    \textbf{User Query:} What type of vehicles did the European Parliament approve a ban on selling from 2035 on February 14, 2023?\\
    \textbf{Target Parameters:}\\
    \ttfamily
    \{\\
    \hspace*{1em}"action\_type": "ban", "legislative\_body": "European Parliament",\\
    \hspace*{1em}"subject": "vehicle sales", "decision\_date": "2023-02-14",\\
    \hspace*{1em}"vehicle\_type": "combustion engine"\\
    \}
    \end{tcolorbox}
    
    % \vspace{0.5em}

    % --- Non-Thinking Mode ---
    \begin{tcolorbox}[colback=green!2, colframe=green!40!black, title=\textbf{Non-Thinking Mode}, arc=1mm, boxrule=0.5mm, fontupper=\ttfamily\scriptsize, breakable]
[\\
  \hspace*{1em}\textcolor{gray}{// Turn 1: Misses the optional `vehicle\_type' parameter}\\
  \hspace*{1em}\{"role": "assistant",\\
    \hspace*{2em}"tool\_calls": [\{"name": "legislative\_action\_retriever",\\
    \hspace*{2em}"arguments": \{"action\_type": "ban", "subject": "vehicle sales", ...\}\}]\},\\
  \hspace*{1em}\{"role": "tool", "content": "No specific legislative action found..."\},\\
  ~\\
  \hspace*{1em}\textcolor{gray}{// Turn 2: Retains previous parameters and successfully isolates the missing variable}\\
  \hspace*{1em}\{"role": "assistant",\\
    \hspace*{2em}"tool\_calls": [\{\\
      \hspace*{3em}"name": "legislative\_action\_retriever",\\
      \hspace*{3em}"arguments": \{"action\_type": "ban", "subject": "vehicle sales", \textcolor{green!60!black}{"vehicle\_type": "combustion engine"}, ...\}\}]\},\\
  \hspace*{1em}\{"role": "tool", "content": "The European Parliament approved a ban..."\}\\
]
    \end{tcolorbox}

    % \vspace{0.5em}
    
    % --- Thinking Mode ---
    \begin{tcolorbox}[colback=red!2, colframe=red!50!black, title=\textbf{Thinking Mode}, arc=1mm, boxrule=0.5mm, fontupper=\ttfamily\scriptsize, breakable]
[\\
  \hspace*{1em}\textcolor{gray}{// Turn 1: Misses the optional `vehicle\_type' parameter}\\
  \hspace*{1em}\{"role": "assistant",\\
    \hspace*{2em}"tool\_calls": [\{"name": "legislative\_action\_retriever",\\
    \hspace*{2em}"arguments": \{"action\_type": "ban", "subject": "vehicle sales", ...\}\}]\},\\
  \hspace*{1em}\{"role": "tool", "content": "No specific legislative action found..."\},\\
  ~\\
  \hspace*{1em}\textcolor{red}{// Turn 2: Overthinks and modifies multiple parameters simultaneously}\\
  \hspace*{1em}\{"role": "assistant",\\
    \hspace*{2em}"content": "[REFLECTION] The tool returned no results. Let me try adjusting the parameters — I'll broaden the subject and also try specifying the vehicle type.",\\
    \hspace*{2em}"tool\_calls": [\{\\
      \hspace*{3em}"name": "legislative\_action\_retriever",\\
      \hspace*{3em}"arguments": \{\\
        \hspace*{4em}\textcolor{red}{"subject": "emissions"}, \textcolor{red}{\textit{// <--- Broken: changed from correct value}}\\
        \hspace*{4em}\textcolor{green!60!black}{"vehicle\_type": "combustion engine"}, \textcolor{green!60!black}{\textit{// <--- Fixed: successfully added}}\\
        \hspace*{4em}...\}\}]\},\\
  \hspace*{1em}\{"role": "tool", "content": "No specific legislative action found..."\},\\
  ~\\
  \hspace*{1em}\textcolor{red}{// Turn 4: Continued Entanglement (Reverts one parameter, breaks another)}\\
  \hspace*{1em}\{"role": "assistant",\\
    \hspace*{2em}"content": "[REFLECTION] Let me try different subject terms and also try the approval action type, since the question says the Parliament `approved a ban.'",\\
    \hspace*{2em}"tool\_calls": [\{\\
      \hspace*{3em}"name": "legislative\_action\_retriever",\\
      \hspace*{3em}"arguments": \{\\
        \hspace*{4em}\textcolor{red}{"action\_type": "approval"}, \textcolor{red}{\textit{// <--- Broken: changed from correct value}}\\
        \hspace*{4em}\textcolor{green!60!black}{"subject": "vehicle sales"}, \textcolor{green!60!black}{\textit{// <--- Fixed: reverted to correct value}}\\
        \hspace*{4em}...\}\}]\}\\
]
    \end{tcolorbox}

\end{tcolorbox}
\captionof{figure}{A case study on overthinking in Claude Opus 4.6.}
\label{fig:claudeover}
\end{center}

\newpage

\begin{center}
\begin{tcolorbox}[colback=white, colframe=black!70, arc=2mm, boxrule=0.8mm, breakable]

    % --- Task Information ---
    \begin{tcolorbox}[colback=gray!5, colframe=gray!40, title=\textbf{Task Information}, arc=1mm, boxrule=0.5mm]
    \small
    \textbf{User Query:} Where did clashes occur between Palestinians and Israeli police on April 5, 2023?\\
    \textbf{Target Tool Parameters:}\\
    \ttfamily
    \{\\
    \hspace*{1em}"event\_type": "clash",\\
    \hspace*{1em}"date": "2023-04-05",\\
    \hspace*{1em}"location": "Palestine" \textcolor{gray}{\textit{// or "Israel"}}\\
    \}
    \end{tcolorbox}
    
    % \vspace{0.5em}

    % --- Non-Thinking Mode ---
    \begin{tcolorbox}[colback=green!2, colframe=green!40!black, title=\textbf{Non-Thinking Model (Successful Broad Search)}, arc=1mm, boxrule=0.5mm, fontupper=\ttfamily\scriptsize, breakable]
[\\
  \hspace*{1em}\textcolor{gray}{// Turn 1: Uses broad term directly extracted from the prompt}\\
  \hspace*{1em}\{\\
    \hspace*{2em}"role": "assistant",\\
    \hspace*{2em}"tool\_calls": [\{"name": "event\_locator", "arguments": \{"location": \textcolor{green!60!black}{"Israel"}, ...\}\}]\\
  \hspace*{1em}\},\\
  \hspace*{1em}\{"role": "tool", "content": "No events found..."\},\\
  ~\\
  \hspace*{1em}\textcolor{gray}{// Turn 2: Tries another broad regional term, successfully retrieves specific data}\\
  \hspace*{1em}\{\\
    \hspace*{2em}"role": "assistant",\\
    \hspace*{2em}"tool\_calls": [\{"name": "event\_locator", "arguments": \{"location": \textcolor{green!60!black}{"Palestine"}, ...\}\}]\\
  \hspace*{1em}\},\\
  \hspace*{1em}\{"role": "tool", "content": "location: Al-Aqsa Mosque, details: Clashes occurred..."\}\\
]
    \end{tcolorbox}

    % \vspace{0.5em}
    
    % --- Thinking Mode ---
    \begin{tcolorbox}[colback=red!2, colframe=red!50!black, title=\textbf{Thinking Model (Failed Premature Specification)}, arc=1mm, boxrule=0.5mm, fontupper=\ttfamily\scriptsize, breakable]
[\\
  \hspace*{1em}\textcolor{red}{// Turn 1: Overthinks and uses internal priors to guess specific sub-locations}\\
  \hspace*{1em}\{\\
    \hspace*{2em}"role": "assistant",\\
    \hspace*{2em}"content": "[REFLECTION] I need to find where clashes occurred... `Jerusalem' is a likely flashpoint for such clashes, so I'll start there.",\\
    \hspace*{2em}"tool\_calls": [\{"name": "event\_locator", "arguments": \{"location": \textcolor{red}{"Jerusalem"}, ...\}\}]\\
  \hspace*{1em}\},\\
  \hspace*{1em}\{"role": "tool", "content": "No events found..."\},\\
  ~\\
  \hspace*{1em}\textcolor{red}{// Turn 2: Continues guessing specific regions, artificially narrowing search space}\\
  \hspace*{1em}\{\\
    \hspace*{2em}"role": "assistant",\\
    \hspace*{2em}"content": "[REFLECTION] The search for Jerusalem returned no results... The West Bank is another common flashpoint... I'll search there.",\\
    \hspace*{2em}"tool\_calls": [\{"name": "event\_locator", "arguments": \{"location": \textcolor{red}{"West Bank"}, ...\}\}]\\
  \hspace*{1em}\},\\
  \hspace*{1em}\{"role": "tool", "content": "No events found..."\}\\
  \hspace*{1em}\textcolor{gray}{// Tool budget exhausted. Strategy Failed.}\\
]
    \end{tcolorbox}

\end{tcolorbox}
\captionof{figure}{A case study on overthinking in Kimi K2.5.}
\label{fig:kimiover}
\end{center}

\newpage

\begin{center}
\begin{tcolorbox}[colback=white, colframe=black!70, arc=2mm, boxrule=0.8mm, breakable]

    % --- Task Information ---
    \begin{tcolorbox}[colback=gray!5, colframe=gray!40, title=\textbf{Task Information}, arc=1mm, boxrule=0.5mm]
    \small
    \textbf{User Query:} Which is greater: (a) the number of species in the forest researched by the biologist who published a paper on rainforest ecosystems, or (b) the weight in kilograms of the satellite launched by the space agency in the country known for its innovation in electronics?
    \end{tcolorbox}
    
    % \vspace{0.5em}

    % --- Non-Thinking Mode ---
    \begin{tcolorbox}[colback=green!2, colframe=green!40!black, title=\textbf{Non-Thinking Model (Corrects behavior based on feedback)}, arc=1mm, boxrule=0.5mm, fontupper=\ttfamily\scriptsize, breakable]
[\\
  \hspace*{1em}\textcolor{gray}{// Turn 1: Makes a single tool call, violating the parallel constraint}\\
  \hspace*{1em}\{\\
    \hspace*{2em}"role": "assistant",\\
    \hspace*{2em}"tool\_calls": [\{"name": "academic\_paper\_finder", ...\}]\\
  \hspace*{1em}\},\\
  \hspace*{1em}\{\\
    \hspace*{2em}"role": "tool", \\
    \hspace*{2em}"content": "ERROR: Tool `academic\_paper\_finder' parallel requirement not met: should be called in parallel with one of [academic\_paper\_finder, innovation\_country\_identifier]."\\
  \hspace*{1em}\},\\
  ~\\
  \hspace*{1em}\textcolor{gray}{// Turn 2: Reads the error message and complies by calling both simultaneously}\\
  \hspace*{1em}\{\\
    \hspace*{2em}"role": "assistant",\\
    \hspace*{2em}"tool\_calls": [\\
      \hspace*{3em}\{"name": \textcolor{green!60!black}{"academic\_paper\_finder"}, ...\},\\
      \hspace*{3em}\{"name": \textcolor{green!60!black}{"innovation\_country\_identifier"}, ...\}\\
    \hspace*{2em}]\\
  \hspace*{1em}\},\\
  \hspace*{1em}\{"role": "tool", "content": "Dr. Jane Goodall published a paper..."\}\\
]
    \end{tcolorbox}

    % \vspace{0.5em}
    
    % --- Thinking Mode ---
    \begin{tcolorbox}[colback=red!2, colframe=red!50!black, title=\textbf{Thinking Model (State Hallucination ignores feedback)}, arc=1mm, boxrule=0.5mm, fontupper=\ttfamily\scriptsize, breakable]
[\\
  \hspace*{1em}\textcolor{gray}{// Previous Turn: Makes an invalid parallel call grouping}\\
  \hspace*{1em}\{\\
    \hspace*{2em}"role": "assistant",\\
    \hspace*{2em}"tool\_calls": [\\
      \hspace*{3em}\{"name": "academic\_paper\_finder"\},\\
      \hspace*{3em}\{"name": "satellite\_info\_retriever"\} \textcolor{gray}{\textit{// Invalid pairing}}\\
    \hspace*{2em}]\\
  \hspace*{1em}\},\\
  \hspace*{1em}\{\\
    \hspace*{2em}"role": "tool", \\
    \hspace*{2em}"content": "ERROR: Tool `academic\_paper\_finder' parallel requirement not met: should be called in parallel with one of [academic\_paper\_finder, innovation\_country\_identifier]."\\
  \hspace*{1em}\},\\
  ~\\
  \hspace*{1em}\textcolor{red}{// Current Turn: Hallucinates past state to rationalize ignoring the error constraint}\\
  \hspace*{1em}\{\\
    \hspace*{2em}"role": "assistant",\\
    \hspace*{2em}"content": "[REFLECTION] The error says academic\_paper\_finder can only be parallel... \textcolor{red}{but we already did innovation\_country\_identifier, got Japan, so maybe first just re-run the academic\_paper\_finder alone now...}",\\
    \hspace*{2em}"tool\_calls": [\{\\
      \hspace*{3em}"name": \textcolor{red}{"academic\_paper\_finder"} \textcolor{red}{\textit{// Fails again by calling it alone}}\\
    \hspace*{2em}\}]\\
  \hspace*{1em}\},\\
  \hspace*{1em}\{\\
    \hspace*{2em}"role": "tool", \\
    \hspace*{2em}"content": "ERROR: Tool `academic\_paper\_finder' parallel requirement not met..."\\
  \hspace*{1em}\}\\
]
    \end{tcolorbox}

\end{tcolorbox}
\captionof{figure}{A case study on how Seed-2.0-Pro becomes more confident in its incorrect solution.}
\label{fig:seedself}
\end{center}

\newpage

\begin{center}
    \begin{tcolorbox}[colback=gray!5, colframe=gray!50, title=\textbf{Prompt Template for Controlled Constraint Expansion}, arc=1mm, boxrule=0.5mm, breakable]
    \small
    You are an expert Constraint Generator for evaluation datasets. Your task is to analyze a specific User Query and a provided Toolset, and then generate a set of rigorous, testable constraints that an AI Agent must follow while solving the problem.
    
    \vspace{1em}
    {\large \textbf{Target Agent Profile}}
    \vspace{0.2em}\hrule\vspace{0.5em}
    To generate appropriate constraints, you must understand the capabilities and behavior of the Target Agent that will execute these tasks:
    \begin{enumerate}
        \setlength{\itemsep}{0pt}
        \setlength{\parskip}{0pt}
        \item \textbf{Tool-Driven Knowledge:} The Agent cannot rely on internal parametric knowledge to answer questions directly. All information must be obtained by invoking the provided tools.
        \item \textbf{Resilience \& Self-Correction:} The Agent operates in a loop. If a tool call fails or returns an error, the Agent is expected to analyze the error, correct its parameters, and retry until success.
        \item \textbf{Parallel Execution:} The Agent is capable of Parallel Calling. It can invoke multiple different tools within a single Action turn to improve efficiency.
    \end{enumerate}
    
    \vspace{1em}
    {\large \textbf{Main Directive}}
    \vspace{0.2em}\hrule\vspace{0.5em}
    You must carefully analyze and select specific constraints from the [New Constraint List]. Then, based on the original question in the provided [Data], the available tools in the [Toolset] and the correct example trajectory [Correct Trajectory List], generate new data that adheres to the requirements below.
    
    \vspace{1em}
    {\large \textbf{Data Generation Requirements}}
    \vspace{0.2em}\hrule\vspace{0.5em}
    \begin{enumerate}
        \setlength{\itemsep}{0pt}
        \setlength{\parskip}{0pt}
        \item Ensure only \texttt{\{c1\}} added, that is, \texttt{\{c2\}}. The word following [Main Category] should be the main category.
        \item Based on this analysis, select \texttt{\{c3\}} from the [New Constraint List] and construct an appropriate "Specific Constraint Content". Add it to the [Original Constraint List], and return the [Updated Constraint List].
        \item Modify the content of the [Original Question] to explicitly and clearly specify all the constraints in the [Updated Constraint List]. The modified question must clearly describe each constraint in natural language.
        \item Ensure that the Specific Constraint in each constraint triplet is detailed and specific, containing concrete information or examples.
        \item Use the provided [Correct Trajectory List] as a feasibility anchor to ensure your generated constraints are realistic and solvable.
        \begin{itemize}
            \item \textbf{Ensure Compatibility:} The constraints you generate must define a solution space that includes the [Correct Trajectory List] as a valid, compliant path.
            \item \textbf{Avoid Overfitting (CRITICAL):} Do not treat the [Correct Trajectory List] as the only correct answer. You must generalize from this specific example rather than rigidly prescribing its exact steps.
        \end{itemize}
        \item Safeguard the agent's strategic autonomy by actively eliminating any mandates that specify required tools or parameter values.
    \end{enumerate}
    
    \vspace{1em}
    {\large \textbf{Notes}}
    \vspace{0.2em}\hrule\vspace{0.5em}
    \begin{enumerate}
        \setlength{\itemsep}{0pt}
        \setlength{\parskip}{0pt}
        \item The new constraint cannot conflict with the constraints in the [Original Constraint List].
        \item The modified [Question with the New Constraint] must explicitly describe all the constraints in natural language.
        \item Tool-related constraints must only reference the tools available in the [Toolset].
        \item Make sure the Specific Constraint in each constraint triplet is as specific as possible.
        \item Important: The response must strictly follow the [Response Format] exactly as specified.
        \item When generating the modified question, ensure that the language is natural and well-polished.
    \end{enumerate}
    
    \vspace{1em}
    {\large \textbf{Response Format}}
    \vspace{0.2em}\hrule\vspace{0.5em}
    \textbf{[Thinking Process]}: xxx\\
    \textbf{[Updated Constraint List]}: [(Main Category, Subcategory, Specific Constraint), ...]\\
    \textbf{[New Specific Constraint]}: (The newly added specific constraint...)\\
    \textbf{[Question with the New Constraint]}: xxx
    
    \vspace{1em}
    {\large \textbf{Data}}
    \vspace{0.2em}\hrule\vspace{0.5em}
    \textbf{[New Constraint List]}: \texttt{\{new\_constraint\_list\}}\\
    \textbf{[Original Constraint List]}: \texttt{[\{original\_constraint\_list\}]}\\
    \textbf{[Original Question]}: \texttt{\{original\_question\}}\\
    \textbf{[Toolset]}: \texttt{\{toolset\}}\\
    \textbf{[Correct Trajectory List]}: \texttt{\{correct\_trajectory\}}
    \end{tcolorbox}
    \captionof{figure}{Prompt Template for controlled constraint expansion.}
    \label{fig:promptconstraintexpansion}
\end{center}

\begin{center}
    \begin{tcolorbox}[colback=gray!5, colframe=gray!50, title=\textbf{Prompt Template for LLM-Based Filtering}, arc=1mm, boxrule=0.5mm, breakable]
    \small
    You are an expert in constraint validation and logical consistency. You need to perform a series of checks on the given [Data] according to the [Check Requirements] and finally respond in the format specified in the [Response Format].
    
    \vspace{1em}
    {\large \textbf{Check Requirements}}
    \vspace{0.2em}\hrule\vspace{0.5em}
    \begin{enumerate}
        \setlength{\itemsep}{0pt}
        \setlength{\parskip}{0pt}
        \item \textbf{Conflicts Among Constraints:} Check if there are any logical contradictions or mutually exclusive requirements within the "Constraint List" itself (e.g., demanding a JSON format but simultaneously requiring the text to end with a period outside the JSON). Explain your reasoning first and then conclude.
        \item \textbf{Mismatches Between Constraints and Scenario Structures:} Check if the constraints align logically with the "Scenario Structure". For example, ensure no constraint forces an impossible execution path, such as demanding at least 3 interaction turns for a simple "Single-Hop" scenario. Explain your reasoning first and then conclude.
        \item \textbf{Conflicts Between Constraints and Tools:} Check if the constraints contradict the operational realities, parameter limits, or functional requirements of the provided "Tools" necessary to answer the "Question". For example, limiting the agent to 1 tool call when the task strictly requires calling 2 different tools. Explain your reasoning first and then conclude.
    \end{enumerate}
    
    \vspace{1em}
    {\large \textbf{Response Format}}
    \vspace{0.2em}\hrule\vspace{0.5em}
    \textbf{\# 1. Conflicts Among Constraints Check \#}\\
    \textbf{[Specific Explanation]}: [Your detailed explanation here]\\
    \textbf{[Is there any conflict among constraints]}: [Yes/No]
    
    \vspace{0.5em}
    \noindent \textbf{\# 2. Mismatches Between Constraints and Scenario Structures Check \#}\\
    \textbf{[Specific Explanation]}: [Your detailed explanation here]\\
    \textbf{[Is there any mismatch between constraints and the scenario structure]}: [Yes/No]
    
    \vspace{0.5em}
    \noindent \textbf{\# 3. Conflicts Between Constraints and Tools Check \#}\\
    \textbf{[Specific Explanation]}: [Your detailed explanation here]\\
    \textbf{[Is there any conflict between constraints and tools]}: [Yes/No]
    
    \vspace{1em}
    {\large \textbf{Data}}
    \vspace{0.2em}\hrule\vspace{0.5em}
    \textbf{[Question]}: \texttt{\{question\}}\\
    \textbf{[Scenario Structure]}: \texttt{\{scenario\_structure\}}\\
    \textbf{[Constraint List]}: \texttt{\{constraint\_list\}}\\
    \textbf{[Tools]}: \texttt{\{tools\}}
    \end{tcolorbox}
    \captionof{figure}{Prompt template for LLM-based filtering.}
    \label{fig:promptconflictverification}
\end{center}

\newpage

\begin{center}
    \begin{tcolorbox}[colback=gray!5, colframe=gray!50, title=\textbf{Prompt Template for Task Context Integration (Stage I)}, arc=1mm, boxrule=0.5mm, breakable]
    \small
    You are tasked with composing a concise and contextual background for a given task, based on its description and the available toolset. \\
    Your goal is to explain why the task matters, what domain it belongs to, and what general context or motivation surrounds it, without revealing any direct solutions, methods, or answers. \\
    The background should help readers understand the purpose and scope of the task, maintaining a professional and neutral tone. \\
    Ensure the background is coherent, informative, and between 3 to 10 sentences in length.
    
    \vspace{1em}
    {\large \textbf{Steps}}
    \vspace{0.2em}\hrule\vspace{0.5em}
    \begin{enumerate}
        \setlength{\itemsep}{0pt}
        \setlength{\parskip}{0pt}
        \item Carefully analyze the provided [Task Description] and [Toolset] to infer their context and objectives.
        \item Identify the relevant field or problem domain the task belongs to.
        \item Avoid including any hints, examples, or methodological guidance for solving the task.
        \item Compose a natural, domain-relevant background paragraph that introduces the task's context and importance.
        \item Present your reasoning process and the resulting background strictly following the format below.
    \end{enumerate}
    
    \vspace{1em}
    {\large \textbf{Output Format}}
    \vspace{0.2em}\hrule\vspace{0.5em}
    Use the following structure in your response (do not use JSON):
    
    \vspace{0.5em}
    \noindent \textbf{[Think]}: <Briefly explain your reasoning and how you derived the background.> \\
    \textbf{[Background]}: <Write a 3–10 sentence background paragraph relevant to the task and toolset.>
    
    \vspace{1em}
    {\large \textbf{Data}}
    \vspace{0.2em}\hrule\vspace{0.5em}
    \textbf{[Task Description]}: \texttt{\{task\}}\\
    \textbf{[Toolset]}: \texttt{\{toolset\}}
    \end{tcolorbox}
    \captionof{figure}{Prompt template for task context integration (stage I).}
\end{center}

\newpage

\begin{center}
    \begin{tcolorbox}[colback=gray!5, colframe=gray!50, title=\textbf{Prompt Template for Task Context Integration (Stage II)}, arc=1mm, boxrule=0.5mm]
    \small
    You are an expert AI System Architect specializing in "Prompt Engineering". Your goal is to synthesize a specialized Agent System Prompt based on a generic template and specific context provided below.
    
    \vspace{1em}
    {\large \textbf{Transformation Instructions (CRITICAL)}}
    \vspace{0.2em}\hrule\vspace{0.5em}
    You must generate a new System Prompt by modifying the [Generic Base Prompt]. 
    
    \vspace{0.5em}
    \noindent\textbf{CORE OBJECTIVE:} Transform the generic template into a bespoke, domain-specific instruction set. The final result should NOT look like a "fill-in-the-blanks" template. It must feel like it was written from scratch for this specific expert agent, while rigidly strictly enforcing the provided constraints.
    
    \begin{enumerate}
        \setlength{\itemsep}{0pt}
        \setlength{\parskip}{0pt}
        \item \textbf{Persona Adaptation \& Contextual Generalization}:
        \begin{itemize}
            \item \textbf{Domain Abstraction}: Analyze the [Task Background] to identify the general domain.
            \item \textbf{Mission Synthesis}: Extract the methodology and values from the background. Weave them into the agent's core identity.
            \item \textbf{Tone Alignment}: You must rewrite the descriptive text throughout the ENTIRE prompt (including the descriptions inside [THOUGHT] and [ACTION]) to match the professional tone of the domain.
        \end{itemize}
    
        \item \textbf{Constraint Extraction, Mapping, and Injection}:\\
        You must process the [Reference Query \& Constraints] using a strict Extract-Map-Inject protocol.
        \begin{itemize}
            \item \textbf{Step A: Separation}: Remove the specific user question. Extract operational constraints.
            \item \textbf{Step B: Single-Location Mapping}: Assign each extracted constraint to EXACTLY ONE logical home within the execution protocol. Do not repeat constraints across multiple sections.
            \item \textbf{Step C: Verbatim Injection}: Insert constraint logic (numbers, tool names, hard rules) exactly as written. If a specific constraint conflicts with or supersedes a generic sentence in the template, DELETE the generic sentence. Do not keep both.
        \end{itemize}
    
        \item \textbf{Boilerplate Pruning \& Optimization}:
        \begin{itemize}
            \item \textbf{Remove Fluff}: DELETE generic fillers unless they add specific value to the domain task.
            \item \textbf{Condense}: Keep instructions concise. The constraints should be the "stars" of the prompt.
            \item \textbf{Focus}: The final prompt should consist primarily of the domain context and the specific constraints.
        \end{itemize}
    
        \item \textbf{Structure Preservation}:
        \begin{itemize}
            \item You MUST preserve the HEADERS and BLOCK TAGS (\texttt{\# Role}, \texttt{[THOUGHT]}, \texttt{[ACTION]}, \texttt{[REFLECTION]}, \texttt{[FINAL ANSWER]}).
            \item You may rewrite the content under these headers, as long as the underlying logic is preserved.
        \end{itemize}
    \end{enumerate}
    
    \vspace{0.5em}
    \noindent The response must strictly follow the [Response Format] exactly as specified.
    
    \vspace{1em}
    {\large \textbf{Response Format}}
    \vspace{0.2em}\hrule\vspace{0.5em}
    \textbf{[Thinking Process]}: (Explain how you pruned the boilerplate and adapted the tone)\\
    \textbf{[System Prompt]}: (The full, final System Prompt)
    
    \vspace{1em}
    {\large \textbf{Data}}
    \vspace{0.2em}\hrule\vspace{0.5em}
    \textbf{[Generic Base Prompt]}: \texttt{\{general\_prompt\}}\\
    \textbf{[Task Background]}: \texttt{\{task\_background\}}\\
    \textbf{[Reference Query \& Constraints]}: \texttt{\{question\_with\_constraint\}}
    \end{tcolorbox}
    \captionof{figure}{Prompt template for task context integration (stage II).}
    \label{fig:promptcontextintegration}
\end{center}

\newpage

% =========================================================
% 1. Interaction Rounds
% =========================================================
\begin{center}
    \begin{tcolorbox}[colback=gray!5, colframe=gray!50, title=\textbf{Prompt Template for Constraint Validation Code Generation (Interaction Rounds)}, arc=1mm, boxrule=0.5mm, breakable]
    \small
    You are an expert in extracting conditions for instruction-following tasks. Please complete the task according to the following requirements.
    
    \vspace{1em}
    {\large \textbf{Background}}
    \vspace{0.2em}\hrule\vspace{0.5em}
    We have a type of constraint called "interaction rounds constraint", which are used to limit the minimum and maximum number of dialogue rounds for an agent. You need to extract the minimum and maximum number of dialogue rounds based on the user query and the interaction round constraints, and output a JSON object that meets the requirements.
    
    \vspace{1em}
    {\large \textbf{Input}}
    \vspace{0.2em}\hrule\vspace{0.5em}
    \begin{itemize}
        \setlength{\itemsep}{0.5em}
        \item User query (contains interaction rounds constraint and other constraints):\\
        \texttt{========================= User Query Start ============================}\\
        \texttt{\{user\_query\}}\\
        \texttt{========================= User Query End ============================}
        \item Summary of the constraint part in the user query:\\
        \texttt{\{refine\_constraint\}}
        \item The current interaction round constraint to be processed:\\
        \texttt{\{constraint\}}
    \end{itemize}
    
    \vspace{1em}
    {\large \textbf{Your Task}}
    \vspace{0.2em}\hrule\vspace{0.5em}
    \begin{enumerate}
        \setlength{\itemsep}{0pt}
        \setlength{\parskip}{0pt}
        \item Read the user query and the interaction round constraints carefully, and determine the minimum and maximum number of dialogue rounds.
        \item The user query may contain other constraints besides the interaction rounds constraint, but please focus only on the parts related to the interaction round constraints.
        \item If the interaction round constraint does not mention a maximum value, default to "inf"; if it does not mention a minimum value, default to 0.
        \item Output a JSON object according to the output format requirements.
    \end{enumerate}
    
    \vspace{1em}
    {\large \textbf{Output Format}}
    \vspace{0.2em}\hrule\vspace{0.5em}
    You must output only a valid JSON object (do not add escape characters before single quotes inside the JSON), without any additional text. The structure is as follows:\\
    
    \vspace{0.5em}
    \ttfamily
    \{\\
    \quad"min\_round": int, \textcolor{gray}{\textit{// Minimum number of dialogue rounds (0 or other)}}\\
    \quad"max\_round": int or string \textcolor{gray}{\textit{// Maximum number of dialogue rounds ("inf" or other)}}\\
    \}\\
    \normalfont
    \end{tcolorbox}
    \captionof{figure}{Prompt template for constraint validation code generation (interaction rounds).}
    \label{fig:promptinteractionrounds}
\end{center}

\newpage

% =========================================================
% 2. Tool Call Count
% =========================================================
\begin{center}
    \begin{tcolorbox}[colback=gray!5, colframe=gray!50, title=\textbf{Prompt Template for Constraint Validation Code Generation (Tool Call Count)}, arc=1mm, boxrule=0.5mm, breakable]
    \small
    You are an expert in extracting conditions for instruction-following tasks. Please complete the task according to the following requirements.
    
    \vspace{1em}
    {\large \textbf{Background}}
    \vspace{0.2em}\hrule\vspace{0.5em}
    We have a type of constraint called "tool call count constraint", which are used to limit the minimum and maximum total number of tool calls an agent can execute throughout the entire task. You need to extract the minimum and maximum total number of tool calls based on the user query and the tool call count constraint, and output a JSON object that meets the requirements.
    
    \vspace{1em}
    {\large \textbf{Input}}
    \vspace{0.2em}\hrule\vspace{0.5em}
    \begin{itemize}
        \setlength{\itemsep}{0.5em}
        \item User query (contains tool call count constraint and other constraints):\\
        \texttt{========================= User Query Start ============================}\\
        \texttt{\{user\_query\}}\\
        \texttt{========================= User Query End ============================}
        \item Summary of the constraint part in the user query:\\
        \texttt{\{refine\_constraint\}}
        \item The current tool call count constraint to be processed:\\
        \texttt{\{constraint\}}
    \end{itemize}
    
    \vspace{1em}
    {\large \textbf{Your Task}}
    \vspace{0.2em}\hrule\vspace{0.5em}
    \begin{enumerate}
        \setlength{\itemsep}{0pt}
        \setlength{\parskip}{0pt}
        \item Read the user query and the tool call count constraint carefully, and determine the minimum and maximum total number of tool calls the agent needs to execute throughout the entire task.
        \item The user query may contain other constraints besides the tool call count constraint, but please focus only on the parts related to the tool call count constraint.
        \item If the tool call count constraint does not mention a maximum value, default to "inf"; if it does not mention a minimum value, default to 0.
        \item Output a JSON object according to the output format requirements.
    \end{enumerate}
    
    \vspace{1em}
    {\large \textbf{Output Format}}
    \vspace{0.2em}\hrule\vspace{0.5em}
    You must output only a valid JSON object (do not add escape characters before single quotes inside the JSON), without any additional text. The structure is as follows:\\
    
    \vspace{0.5em}
    \ttfamily
    \{\\
    \quad"min\_callTimes": int, \textcolor{gray}{\textit{// Minimum total number of tool calls (0 or other)}}\\
    \quad"max\_callTimes": int or string \textcolor{gray}{\textit{// Maximum total number of tool calls ("inf" or other)}}\\
    \}\\
    \normalfont
    \end{tcolorbox}
    \captionof{figure}{Prompt template for constraint validation code generation (tool call count).}
    \label{fig:prompttoolcallcount}
\end{center}

\newpage

% =========================================================
% 3. Specific Tool Call Count
% =========================================================
\begin{center}
    \begin{tcolorbox}[colback=gray!5, colframe=gray!50, title=\textbf{Prompt Template for Constraint Validation Code Generation (Specific Tool Call Count)}, arc=1mm, boxrule=0.5mm, breakable]
    \small
    You are an expert in extracting conditions for instruction-following tasks. Please complete the task according to the following requirements.
    
    \vspace{1em}
    {\large \textbf{Background}}
    \vspace{0.2em}\hrule\vspace{0.5em}
    We have a type of constraint called "specific tool call count constraint", which are used to limit the maximum number of calls allowed for each specific tool type during the entire task execution process. This constraint sets an independent upper limit for each specified tool. Once a tool reaches its maximum allowed number of calls, the agent is strictly prohibited from using that tool again in the remainder of the task.
    
    \vspace{1em}
    {\large \textbf{Input}}
    \vspace{0.2em}\hrule\vspace{0.5em}
    \begin{itemize}
        \setlength{\itemsep}{0.5em}
        \item User query (contains specific tool call count constraint and other constraints):\\
        \texttt{========================= User Query Start ============================}\\
        \texttt{\{user\_query\}}\\
        \texttt{========================= User Query End ============================}
        \item Summary of the constraint part in the user query:\\
        \texttt{\{refine\_constraint\}}
        \item The current specific tool call count constraint to be processed:\\
        \texttt{\{constraint\}}
        \item List of tools (candidate tool set):\\
        \texttt{\{tools\_name\}}
    \end{itemize}
    
    \vspace{1em}
    {\large \textbf{Your Task}}
    \vspace{0.2em}\hrule\vspace{0.5em}
    \begin{enumerate}
        \setlength{\itemsep}{0pt}
        \setlength{\parskip}{0pt}
        \item Read the user query and the specific tool call count constraint carefully, and extract the maximum call limit for each specified tool.
        \item The user query may contain other constraints besides the specific tool call count constraints, but please focus only on the parts related to the specific tool call count constraints.
        \item Note: This constraint may target multiple different tools, and each tool may have a different limit.
        \item Output a JSON object according to the output format requirements.
    \end{enumerate}
    
    \vspace{1em}
    {\large \textbf{Output Format}}
    \vspace{0.2em}\hrule\vspace{0.5em}
    You must output only a valid JSON object (do not add escape characters before single quotes inside the JSON), without any additional text. The structure is as follows:\\
    
    \vspace{0.5em}
    \ttfamily
    \{\\
    \quad"max\_calls\_per\_tool": \{\\
    \qquad"tool\_name\_1": int, \textcolor{gray}{\textit{// Maximum number of calls for tool 1}}\\
    \qquad"tool\_name\_2": int, \textcolor{gray}{\textit{// Maximum number of calls for tool 2}}\\
    \qquad... \textcolor{gray}{\textit{// possibly more tools}}\\
    \quad\}\\
    \}\\
    \normalfont
    \vspace{0.5em}
    Note: The tool names must be exactly those in the tool list.
    \end{tcolorbox}
    \captionof{figure}{Prompt template for constraint validation code generation (specific tool call count).}
    \label{fig:promptspecifictoolcallcount}
\end{center}

\newpage

% =========================================================
% 4. Sequential Dependencies
% =========================================================
\begin{center}
    \begin{tcolorbox}[colback=gray!5, colframe=gray!50, title=\textbf{Prompt Template for Constraint Validation Code Generation (Sequential Dependencies)}, arc=1mm, boxrule=0.5mm, breakable]
    \small
    You are an expert in extracting conditions for instruction-following tasks. Please complete the task according to the following requirements.
    
    \vspace{1em}
    {\large \textbf{Background}}
    \vspace{0.2em}\hrule\vspace{0.5em}
    We have a type of constraint called "sequential dependencies constraint", which are used to enforce a strict temporal order of tool calls (e.g., tool A must be called before tool B). This constraint focuses on the sequence of tools along the timeline and does not require data or state dependencies between tools.
    
    \vspace{1em}
    {\large \textbf{Input}}
    \vspace{0.2em}\hrule\vspace{0.5em}
    \begin{itemize}
        \setlength{\itemsep}{0.5em}
        \item User query (contains call dependency constraint and other constraints):\\
        \texttt{========================= User Query Start ============================}\\
        \texttt{\{user\_query\}}\\
        \texttt{========================= User Query End ============================}
        \item Summary of the constraint part in the user query:\\
        \texttt{\{refine\_constraint\}}
        \item The current call dependency constraint to be processed:\\
        \texttt{\{constraint\}}
        \item List of tools (candidate tool set):\\
        \texttt{\{tools\_name\}}
    \end{itemize}
    
    \vspace{1em}
    {\large \textbf{Your Task}}
    \vspace{0.2em}\hrule\vspace{0.5em}
    \begin{enumerate}
        \setlength{\itemsep}{0pt}
        \setlength{\parskip}{0pt}
        \item Read the user query and the call dependency constraint carefully, and extract all explicit call dependency relationships.
        \item The user query may contain other constraints besides the call dependency constraint, but please focus only on the parts related to the call dependency constraint.
        \item Identify the sequential relationships between tools mentioned in the constraint (e.g., "A must be before B", "call X first, then Y", etc.).
        \item Represent each sequential relationship as a list of two or more elements: \texttt{[first tool to be called, second tool to be called, ...]}.
        \item If there are multiple sequential relationships, extract all of them.
        \item If there is no clear sequential relationship in the constraint, return an empty list.
        \item Output a JSON object according to the output format requirements.
    \end{enumerate}
    
    \vspace{1em}
    {\large \textbf{Output Format}}
    \vspace{0.2em}\hrule\vspace{0.5em}
    You must output only a valid JSON object (do not add escape characters before single quotes inside the JSON), without any additional text. The structure is as follows:\\
    
    \vspace{0.5em}
    \ttfamily
    \{\\
    \quad"order\_constraints": [\\
    \qquad["tool\_A", "tool\_B", ...], \textcolor{gray}{\textit{// tool\_A, tool\_B, ... have a sequential relationship}}\\
    \qquad... \textcolor{gray}{\textit{// more sequential relationships}}\\
    \quad]\\
    \}\\
    \normalfont
    \vspace{0.5em}
    Note: The tool names must be exactly those in the tool list.
    \end{tcolorbox}
    \captionof{figure}{Prompt template for constraint validation code generation (sequential dependencies).}
    \label{fig:promptcalldependency}
\end{center}

\newpage

% =========================================================
% 5. Parallel Dependencies
% =========================================================
\begin{center}
    \begin{tcolorbox}[colback=gray!5, colframe=gray!50, title=\textbf{Prompt Template for Constraint Validation Code Generation (Parallel Dependencies)}, arc=1mm, boxrule=0.5mm, breakable]
    \small
    You are an expert in extracting conditions for instruction-following tasks. Please complete the task according to the following requirements.
    
    \vspace{1em}
    {\large \textbf{Background}}
    \vspace{0.2em}\hrule\vspace{0.5em}
    We have a type of constraint called "parallel dependencies constraint", which are used to enforce that the agent must call multiple specified tools simultaneously in a single instruction step (e.g., tool A and tool B must be called in the same step). This constraint tests the agent's ability to plan and coordinate parallel processes within a single round.
    
    \vspace{1em}
    {\large \textbf{Input}}
    \vspace{0.2em}\hrule\vspace{0.5em}
    \begin{itemize}
        \setlength{\itemsep}{0.5em}
        \item User query (contains tool-specific parallel calls constraint and other constraints):\\
        \texttt{========================= User Query Start ============================}\\
        \texttt{\{user\_query\}}\\
        \texttt{========================= User Query End ============================}
        \item Summary of the constraint part in the user query:\\
        \texttt{\{refine\_constraint\}}
        \item The current tool-specific parallel calls constraint to be processed:\\
        \texttt{\{constraint\}}
        \item List of tools (candidate tool set):\\
        \texttt{\{tools\_name\}}
    \end{itemize}
    
    \vspace{1em}
    {\large \textbf{Your Task}}
    \vspace{0.2em}\hrule\vspace{0.5em}
    \begin{enumerate}
        \setlength{\itemsep}{0pt}
        \setlength{\parskip}{0pt}
        \item Read the user query and the tool-specific parallel calls constraint carefully, and extract all groups of tools that need to be called simultaneously.
        \item The user query may contain other constraints besides the tool-specific parallel calls constraint, but please focus only on the parts related to the tool-specific parallel calls constraint.
        \item Identify the combinations of tools that need to be called simultaneously as mentioned in the constraint (e.g., "A and B must be called together", "X, Y, and Z must be used in the same step", etc.).
        \item Represent each parallel tool group as a list containing the names of all tools that need to be called simultaneously.
        \item If there are multiple parallel tool groups, extract all of them.
        \item If there is no clear parallel relationship in the constraint, return an empty list.
        \item Output a JSON object according to the output format requirements.
    \end{enumerate}
    
    \vspace{1em}
    {\large \textbf{Output Format}}
    \vspace{0.2em}\hrule\vspace{0.5em}
    You must output only a valid JSON object (do not add escape characters before single quotes inside the JSON), without any additional text. The structure is as follows:\\
    
    \vspace{0.5em}
    \ttfamily
    \{\\
    \quad"parallel\_groups": [\\
    \qquad["tool\_A", "tool\_B", ...], \textcolor{gray}{\textit{// tool\_A, tool\_B and ... must be called simultaneously}}\\
    \qquad["tool\_X", "tool\_Y", "tool\_Z", ...], \textcolor{gray}{\textit{// tool\_X, tool\_Y, tool\_Z and ... must be called simultaneously}}\\
    \qquad... \textcolor{gray}{\textit{// more parallel relationships}}\\
    \quad]\\
    \}\\
    \normalfont
    \vspace{0.5em}
    Note: The tool names must be exactly those in the tool list.
    \end{tcolorbox}
    \captionof{figure}{Prompt template for constraint validation code generation (parallel dependencies).}
    \label{fig:prompttoolspecificparallelcalls}
\end{center}

\newpage

% =========================================================
% 6. Parallel Calls Count
% =========================================================
\begin{center}
    \begin{tcolorbox}[colback=gray!5, colframe=gray!50, title=\textbf{Prompt Template for Constraint Validation Code Generation (Parallel Calls Count)}, arc=1mm, boxrule=0.5mm, breakable]
    \small
    You are an expert in extracting conditions for instruction-following tasks. Please complete the task according to the following requirements.
    
    \vspace{1em}
    {\large \textbf{Background}}
    \vspace{0.2em}\hrule\vspace{0.5em}
    We have a type of constraint called "parallel calls count constraint", which are used to limit the maximum number of tool types or the total number of tools that an agent can call simultaneously in a single interaction round (hereinafter referred to as the maximum), or to limit the minimum number of tool types or the total number of tools that the agent must call simultaneously in at least one round (hereinafter referred to as the minimum). You need to extract the "maximum" and "minimum" values mentioned above based on the user query and the parallel calls count constraint, and also indicate whether they refer to "type" or "number." Then output a JSON object that meets the requirements.
    
    \vspace{1em}
    {\large \textbf{Input}}
    \vspace{0.2em}\hrule\vspace{0.5em}
    \begin{itemize}
        \setlength{\itemsep}{0.5em}
        \item User query (contains parallel calls count constraint and other constraints):\\
        \texttt{========================= User Query Start ============================}\\
        \texttt{\{user\_query\}}\\
        \texttt{========================= User Query End ============================}
        \item Summary of the constraint part in the user query:\\
        \texttt{\{refine\_constraint\}}
        \item The current parallel calls count constraint to be processed:\\
        \texttt{\{constraint\}}
    \end{itemize}
    
    \vspace{1em}
    {\large \textbf{Your Task}}
    \vspace{0.2em}\hrule\vspace{0.5em}
    \begin{enumerate}
        \setlength{\itemsep}{0pt}
        \setlength{\parskip}{0pt}
        \item Read the user query and the parallel calls count constraint carefully, and determine the maximum number of different tool types that the agent can call simultaneously in a single interaction round (maximum) and the minimum number of different tool types that the agent must call simultaneously in at least one round (minimum).
        \item The user query may contain other constraints besides the parallel calls count constraint, but please focus only on the parts related to the parallel calls count constraint.
        \item Pay attention to whether the parallel calls count constraint refers to tool types or the total number of tools: it may be "type" or "num". If not explicitly specified, default to "type".
        \item If the parallel calls count constraint does not mention a maximum value, default to "inf"; if it does not mention a minimum value, default to 0.
        \item Output a JSON object according to the output format requirements.
    \end{enumerate}
    
    \vspace{1em}
    {\large \textbf{Output Format}}
    \vspace{0.2em}\hrule\vspace{0.5em}
    You must output only a valid JSON object (do not add escape characters before single quotes inside the JSON), without any additional text. The structure is as follows:\\
    
    \vspace{0.5em}
    \ttfamily
    \{\\
    \quad"min\_parallelCallTypes": int, \textcolor{gray}{\textit{// Minimum value (0 or other)}}\\
    \quad"max\_parallelCallTypes": int or string, \textcolor{gray}{\textit{// Maximum value ("inf" or other)}}\\
    \quad"unit": string \textcolor{gray}{\textit{// Unit: "type" or "num"}}\\
    \}\\
    \normalfont
    \end{tcolorbox}
    \captionof{figure}{Prompt template for constraint validation code generation (parallel calls count).}
    \label{fig:promptparallelcallscount}
\end{center}

\newpage

% =========================================================
% Validation Code 1: Available Tools and Parameters
% =========================================================
\begin{center}
    \begin{tcolorbox}[colback=gray!5, colframe=gray!50, title=\textbf{Validation Code Snippet (Available Tools and Parameters)}, arc=1mm, boxrule=0.5mm, breakable]
    \small
    \vspace{0.5em}
    \ttfamily
    \textcolor{gray}{\# Check 1: Tool Existence}\\
    if name not in self.tools\_doc:\\
    \quad err = f"Failed to call tool `\{name\}' as it does not exist"\\
    ~\\
    \textcolor{gray}{\# Check 2: Hallucinated Parameters}\\
    args\_keys = set(args.keys())\\
    extra = sorted(args\_keys - tool\_doc\_keys)\\
    if extra:\\
    \quad err = f"Failed to call tool `\{name\}' due to extra argument(s): \{`, '.join(extra)\}"\\
    \normalfont
    \vspace{0.5em}
    \end{tcolorbox}
    \captionof{figure}{Validation code snippet for explicitly enforcing available tools and parameters constraints.}
    \label{fig:codeavailabletools}
\end{center}

% =========================================================
% Validation Code 2: Required Parameters
% =========================================================
\begin{center}
    \begin{tcolorbox}[colback=gray!5, colframe=gray!50, title=\textbf{Validation Code Snippet (Required Parameters)}, arc=1mm, boxrule=0.5mm, breakable]
    \small
    \vspace{0.5em}
    \ttfamily
    \textcolor{gray}{\# Check: Missing Required Arguments}\\
    args\_keys = set(args.keys())\\
    missing = [p for p in required if p not in args\_keys]\\
    if missing:\\
    \quad err = f"Failed to call tool `\{name\}' due to missing required argument(s): \{`,'.join(missing)\}"\\
    \normalfont
    \vspace{0.5em}
    \end{tcolorbox}
    \captionof{figure}{Validation code snippet for explicitly enforcing required parameters constraints.}
    \label{fig:coderequiredparameters}
\end{center}

% =========================================================
% Validation Code 3: Parameter Types
% =========================================================
\begin{center}
    \begin{tcolorbox}[colback=gray!5, colframe=gray!50, title=\textbf{Validation Code Snippet (Parameter Types)}, arc=1mm, boxrule=0.5mm, breakable]
    \small
    \vspace{0.5em}
    \ttfamily
    \textcolor{gray}{\# Check: JSON Schema Type Validation}\\
    expected\_type = schema.get("type")\\
    if expected\_type is not None and (not \_value\_matches\_json\_type(value, expected\_type)):\\
    \quad errors.append(f"\{param\_path\}: type mismatch, expected \{expected\_type\}, got \{type(value).\_\_name\_\_\}")\\
    \quad return errors\\
    \normalfont
    \vspace{0.5em}
    \end{tcolorbox}
    \captionof{figure}{Validation code snippet for explicitly enforcing parameter types constraints through recursive schema validation.}
    \label{fig:codeparametertypes}
\end{center}

\newpage

% =========================================================
% 7. Response Length
% =========================================================
\begin{center}
    \begin{tcolorbox}[colback=gray!5, colframe=gray!50, title=\textbf{Prompt Template for Constraint Validation Code Generation (Length)}, arc=1mm, boxrule=0.5mm, breakable]
    \small
    You are an expert in extracting conditions for instruction-following tasks. Please complete the task according to the following requirements.
    
    \vspace{1em}
    {\large \textbf{Background}}
    \vspace{0.2em}\hrule\vspace{0.5em}
    We have a type of constraint called "response length constraint", which are used to limit the length of the agent's entire response (calculated by word count or character count). This constraint ensures the conciseness, relevance, or level of detail of the response.
    
    \vspace{1em}
    {\large \textbf{Input}}
    \vspace{0.2em}\hrule\vspace{0.5em}
    \begin{itemize}
        \setlength{\itemsep}{0.5em}
        \item User query (contains response length constraint and other constraints):\\
        \texttt{========================= User Query Start ============================}\\
        \texttt{\{user\_query\}}\\
        \texttt{========================= User Query End ============================}
        \item Summary of the constraint part in the user query:\\
        \texttt{\{refine\_constraint\}}
        \item The current response length constraint to be processed:\\
        \texttt{\{constraint\}}
    \end{itemize}
    
    \vspace{1em}
    {\large \textbf{Your Task}}
    \vspace{0.2em}\hrule\vspace{0.5em}
    \begin{enumerate}
        \setlength{\itemsep}{0pt}
        \setlength{\parskip}{0pt}
        \item Read the user query and the response length constraint carefully, and determine the minimum and maximum length of the agent's response.
        \item The user query may contain other constraints besides the response length constraint, but please focus only on the parts related to the response length constraint.
        \item Pay attention to the unit of length: it may be words or characters. If not explicitly specified, default to "characters".
        \item If the response length constraint does not mention a maximum value, default to "inf"; if it does not mention a minimum value, default to 0.
        \item Output a JSON object according to the output format requirements.
    \end{enumerate}
    
    \vspace{1em}
    {\large \textbf{Output Format}}
    \vspace{0.2em}\hrule\vspace{0.5em}
    You must output only a valid JSON object (do not add escape characters before single quotes inside the JSON), without any additional text. The structure is as follows:\\
    
    \vspace{0.5em}
    \ttfamily
    \{\\
    \quad"min\_responseLength": int, \textcolor{gray}{\textit{// Minimum response length (0 or other)}}\\
    \quad"max\_responseLength": int or string, \textcolor{gray}{\textit{// Maximum response length ("inf" or other)}}\\
    \quad"unit": string \textcolor{gray}{\textit{// Length unit: "words" or "characters"}}\\
    \}\\
    \normalfont
    \end{tcolorbox}
    \captionof{figure}{Prompt template for constraint validation code generation (length).}
    \label{fig:promptresponselength}
\end{center}

\newpage

% =========================================================
% 8. Format & Content
% =========================================================
\begin{center}
    \begin{tcolorbox}[colback=gray!5, colframe=gray!50, title=\textbf{Prompt Template for Constraint Validation Code Generation (Format \& Content)}, arc=1mm, boxrule=0.5mm, breakable]
    \small
    You are a code generation expert. Please generate Python validator code according to the following requirements.
    
    \vspace{1em}
    {\large \textbf{Background}}
    \vspace{0.2em}\hrule\vspace{0.5em}
    We have a type of constraint called "response format and content constraint", which includes the following secondary constraints:
    \begin{itemize}
        \setlength{\itemsep}{0pt}
        \setlength{\parskip}{0pt}
        \item Response format constraint (format)
        \item Content constraint (content)
    \end{itemize}
    Your task is: for multiple secondary constraints appearing in the same user query, generate corresponding multiple Python validator functions at once, one function per constraint.
    
    \vspace{1em}
    {\large \textbf{Input}}
    \vspace{0.2em}\hrule\vspace{0.5em}
    \begin{itemize}
        \setlength{\itemsep}{0.5em}
        \item User query (contains response format and content constraint and other constraints):\\
        \texttt{========================= User Query Start ============================}\\
        \texttt{\{user\_query\}}\\
        \texttt{========================= User Query End ============================}
        \item Summary of the constraint part in the user query:\\
        \texttt{\{refine\_constraint\}}
        \item List of secondary constraints to be processed currently.\\
        \texttt{\{constraints\}}
    \end{itemize}
    
    \vspace{1em}
    {\large \textbf{Your Task}}
    \vspace{0.2em}\hrule\vspace{0.5em}
    \begin{enumerate}
        \setlength{\itemsep}{0pt}
        \setlength{\parskip}{0pt}
        \item Read the user query and the list of response format and content constraint mentioned above, but focus only on the parts related to these response format and content constraint.
        \item For each type of response format and content constraint that appears, generate a separate validator function (the function name must strictly follow the rules below):
        \begin{itemize}
            \item format $\rightarrow$ \texttt{validate\_format(response: str) -> Tuple[bool, str]}
            \item content $\rightarrow$ \texttt{validate\_content(response: str) -> Tuple[bool, str]}
        \end{itemize}
        If a certain type does not appear, there is no need to generate the corresponding function.
        \item Each function must:
        \begin{itemize}
            \item Receive \texttt{response: str}
            \item Return \texttt{(bool, str)}, where the \texttt{str} must be in English and sufficiently detailed so that the agent can correctly modify the output based on it.
        \end{itemize}
        \item The code can share auxiliary functions, regular expressions, parsing logic, but different constraints must be distinguished by different functions.
        \item Your code should include necessary imports and comments.
        \item Only output the code, do not output any additional explanation.
    \end{enumerate}
    
    \vspace{1em}
    {\large \textbf{Output Format}}
    \vspace{0.2em}\hrule\vspace{0.5em}
    You must output only a Python code block (enclosed in triple backticks and python: \texttt{\`{}\`{}\`{}python}), without any extra text.\\
    The code block should contain necessary imports, optional auxiliary functions, and several \texttt{validate\_*} functions.
    \end{tcolorbox}
    \captionof{figure}{Prompt template for constraint validation code generation (format \& content).}
    \label{fig:promptformatcontent}
\end{center}

\end{document}